\documentclass[10pt,twocolumn,letterpaper]{article}

\usepackage{cvpr}
\usepackage{times}
\usepackage{epsfig}
\usepackage{graphicx}
\usepackage{amsmath}
\usepackage{amssymb}
\usepackage{multirow}
\usepackage{url}
\usepackage{wrapfig}
\usepackage{subfigure}
\usepackage{array}

\usepackage[pagebackref=true,breaklinks=true,letterpaper=true,colorlinks,bookmarks=false]{hyperref}

\cvprfinalcopy 

\def\cvprPaperID{1288} 

\ifcvprfinal\fi

\begin{document}

\title{PointRNN: Point Recurrent Neural Network for Moving Point Cloud Processing}

\author{Hehe Fan, Yi Yang\\
University of Technology Sydney
}

\maketitle

\begin{abstract}
In this paper, we introduce a Point Recurrent Neural Network (PointRNN) for moving point cloud processing. 
At each time step, PointRNN takes point coordinates $\boldsymbol{P} \in \mathbb{R}^{n \times 3}$ and point features $\boldsymbol{X} \in \mathbb{R}^{n \times d}$ as input ($n$ and $d$ denote the number of points and the number of feature channels, respectively). 
The state of PointRNN is composed of point coordinates $\boldsymbol{P}$ and point states $\boldsymbol{S} \in \mathbb{R}^{n \times d'}$ ($d'$ denotes the number of state channels).
Similarly, the output of PointRNN is composed of $\boldsymbol{P}$ and new point features $\boldsymbol{Y} \in \mathbb{R}^{n \times d''}$ ($d''$ denotes the number of new feature  channels).
Since point clouds are orderless, point features and states from two time steps can not be directly operated. 
Therefore, a point-based spatiotemporally-local correlation is adopted to aggregate point features and states according to point coordinates.
We further propose two variants of PointRNN, \ie, Point Gated Recurrent Unit (PointGRU) and Point Long Short-Term Memory (PointLSTM).
We apply PointRNN, PointGRU and PointLSTM to moving point cloud prediction, which aims to predict the future trajectories of points in a set given their history movements.
Experimental results show that PointRNN, PointGRU and PointLSTM are able to produce correct predictions on both synthetic and real-world datasets, demonstrating their ability to model point cloud sequences. 
The code has been released at \url{https://github.com/hehefan/PointRNN}.
\end{abstract}

\section{Introduction}

Most modern robot and self-driving car platforms rely on 3D point clouds for geometry perception.
In contrast to RGB images, point clouds can provide accurate displacement measurements, which are generally unaffected by lighting conditions. 
Therefore, point cloud is attracting more and more attention in the community. 
However, most of the existing works focus on static point cloud analysis, \eg, classification, segmentation and detection~\cite{DBLP:conf/cvpr/QiSMG17,DBLP:conf/nips/QiYSG17,DBLP:conf/nips/LiBSWDC18,DBLP:journals/iccv/abs-1904-09664}.
Few works study dynamic point clouds.
Intelligent systems need the ability to understand not only the static scenes around them but also the dynamic changes in the environment.
In this paper, we propose a Point Recurrent Neural Network (PointRNN) and its two variants for moving point cloud processing.

In the general setting, a point cloud is a set of points in 3D space.
Usually, the point cloud is represented by the three coordinates $\boldsymbol{P} \in \mathbb{R}^{n \times 3}$ of points and their features $\boldsymbol{X} \in \mathbb{R}^{n \times d}$ (if features are provided), where $n$ and $d$ denote the number of points and feature channels, respectively.
Essentially, point clouds are unordered sets and invariant to permutations of their points.
For example, the set $\{(1,1,1), (2,2,2), (3,3,3)\}$ and $\{(3,3,3), (1,1,1), (2,2,2)\}$ represent the same point cloud.
This irregular format data structure considerably increases the challenge to reason point clouds, making many existing achievements of deep neural networks on image and video fail to directly process point clouds.
\begin{figure}[t]
\footnotesize
\centering
\includegraphics[width=0.985\linewidth]{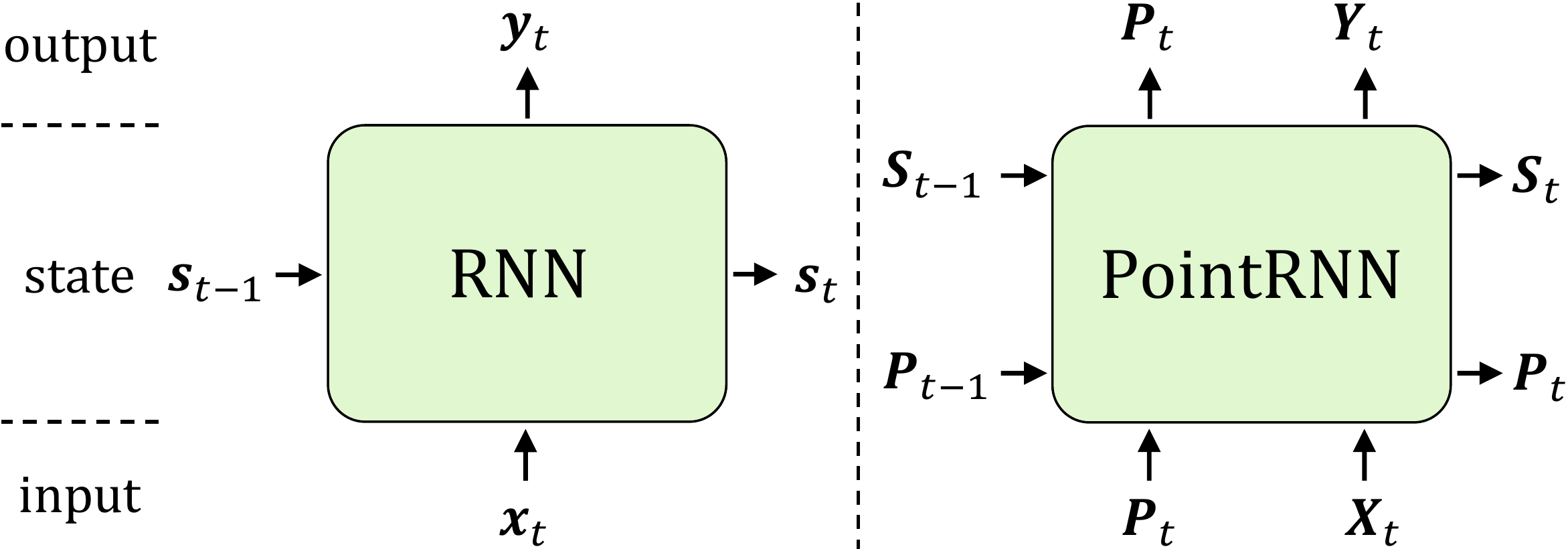}
\caption{Comparison between RNN and PointRNN. 
At the time step $t$, RNN takes a vector $\boldsymbol{x}_t$ as input, updates its state $\boldsymbol{s}_{t-1}$ to $\boldsymbol{s}_{t}$ and outputs a vector $\boldsymbol{y}_t$. 
PointRNN takes point coordinates $\boldsymbol{P}_{t}$ and point features $\boldsymbol{X}_{t}$ as inputs, updates its state $(\boldsymbol{P}_{t-1}, \boldsymbol{S}_{t-1})$ to $(\boldsymbol{P}_t, \boldsymbol{S}_t)$, and outputs $\boldsymbol{P}_{t}$ and new point features $\boldsymbol{Y}_{t}$. 
}
\label{fig:intro}
\end{figure}

The recurrent neural network (RNN) and its variants, \eg, Long Short-Term Memory (LSTM)~\cite{DBLP:journals/neco/HochreiterS97} and Gated Recurrent Unit (GRU)~\cite{DBLP:conf/emnlp/ChoMGBBSB14}, are well-suited to processing time series data.
Generally, the (vanilla) RNN looks at a vector $\boldsymbol{x}_t$ at the time step $t$, updates its internal state (memory) $\boldsymbol{s}_{t-1}$ to $\boldsymbol{s}_t$, and outputs a new vector  $\boldsymbol{y}_t$.
The behavior of RNN can be formulated as follows,
\begin{align*}
\small
\boldsymbol{y}_t, \boldsymbol{s}_t =  \mathrm{RNN}(\boldsymbol{x}_t, \boldsymbol{s}_{t-1}; \boldsymbol{\theta}),
\end{align*}
where $\boldsymbol{\theta}$ denotes the parameters of RNN.
Since moving point cloud is a kind of temporal sequence, we can exploit RNN to process it.

However, the conventional RNN has two severe limitations on processing point cloud sequences.
On one hand, RNN learns from one-dimensional vectors, in which the input, state and output are highly compact.
It is difficult for a vector to represent an entire point cloud.
Although the $\boldsymbol{P}$ and $\boldsymbol{X}$ can be flatten to one-dimensional vectors, such operation heavily damages the data structure and increases the challenge for neural networks to understand point clouds.
Moreover, this operation is not independent of point permutations.
If we use the global feature to represent a point cloud, the local structure will be lost.
To overcome this problem, PointRNN directly takes $(\boldsymbol{P}, \boldsymbol{X})$ as input.
Similarly, the one-dimensional state $\boldsymbol{s}$ and output $\boldsymbol{y}$ in RNN are extended to two-dimensional $\boldsymbol{S} \in \mathbb{R}^{n \times d'}$ and $\boldsymbol{Y} \in \mathbb{R}^{n \times d''}$ in PointRNN, in which each row corresponds to a point.
Besides, because $\boldsymbol{S}$ and $\boldsymbol{Y}$ depends on point coordinates,
$\boldsymbol{P}$ is added into the state and output of PointRNN.
The behavior of PointRNN can be formulated as follows, 
\begin{align*}
\footnotesize
(\boldsymbol{P}_t, \boldsymbol{Y}_t), (\boldsymbol{P}_t, \boldsymbol{S}_t)  =  \mathrm{PointRNN}\big((\boldsymbol{P}_t, \boldsymbol{X}_t), (\boldsymbol{P}_{t-1}, \boldsymbol{S}_{t-1}); \boldsymbol{\theta}\big).
\end{align*}
We illustrate an RNN and a PointRNN unit in Figure~\ref{fig:intro}.

\begin{figure*}[t]
\centering
\includegraphics[width=0.985\linewidth]{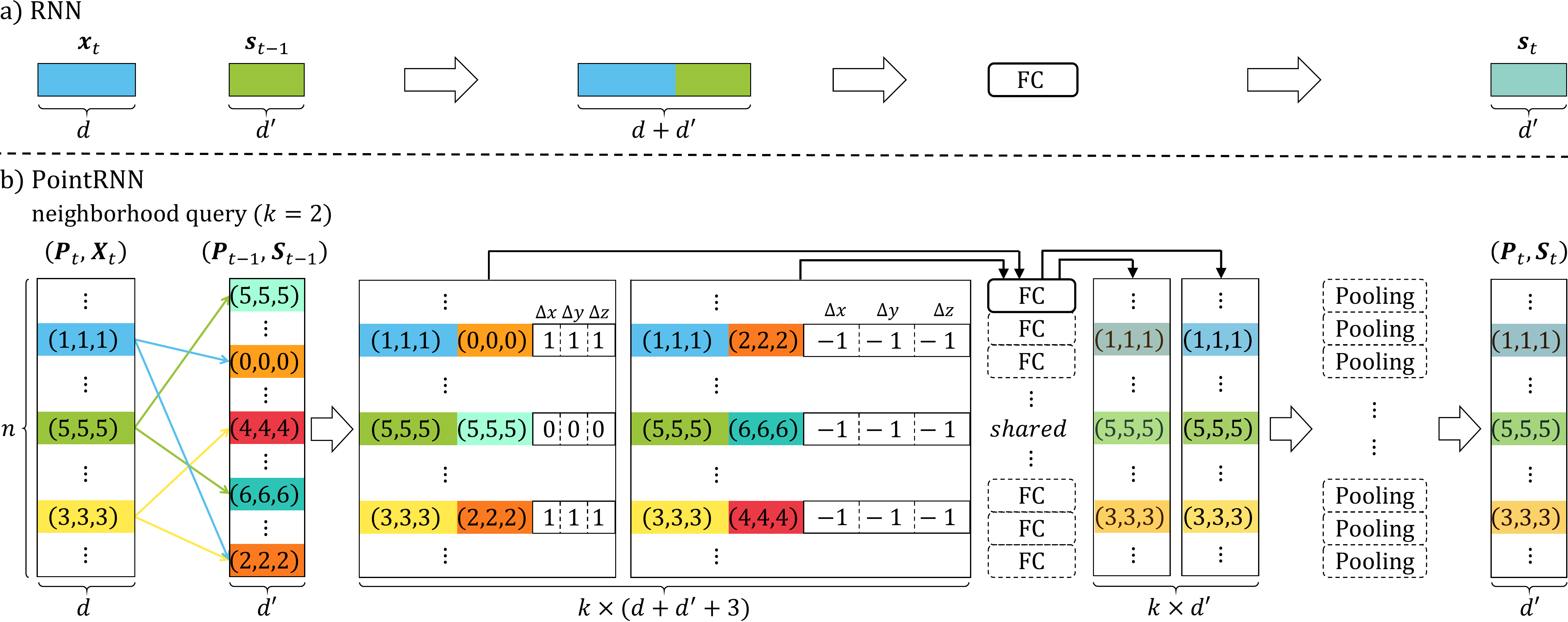}
\footnotesize
\caption{a) The RNN aggregates the input $\boldsymbol{x}_t \in \mathbb{R}^d$ and state $\boldsymbol{s}_{t-1} \in \mathbb{R}^{d'}$ by a concatenation operation, followed by an fully-connected (FC) layer. 
b) The PointRNN aggregates $\boldsymbol{X}_t\in \mathbb{R}^{n\times d}$ and $\boldsymbol{S}_{t-1} \in \mathbb{R}^{n \times d'}$ according to point coordinates by a spatiotemporally-local correlation.  
First, for each point in $\boldsymbol{P}_{t} \in \mathbb{R}^{n \times 3}$, PointRNN searches $k$ neighbors in $\boldsymbol{P}_{t-1}$.
Second, for each neighbor, the feature of the query point in $\boldsymbol{X}_t$, the state of the neighbor in $\boldsymbol{S}_{t-1}$, and the displacement from the neighbor to the query point are concatenated.
Third, the $k$ concatenated representations are processed by an FC layer. 
At last, pooling is used to merge the $k$ representations.
}
\label{fig:unit}
\end{figure*}

One the other hand, RNN aggregates the previous state $\boldsymbol{s}_{t-1}$ and the current input $\boldsymbol{x}_t$ based on a concatenation operation (Figure~\ref{fig:unit}(a)).
However, because point clouds are unordered, concatenation can not be directly applied to point clouds. 
To solve this problem, we adopt a spatiotemporally-local correlation to aggregate $\boldsymbol{X}_t$ and $\boldsymbol{S}_{t-1}$ according to point coordinates (Figure~\ref{fig:unit}(b)).
Specifically, for each point in $\boldsymbol{P}_{t}$, PointRNN first searches its neighbors in $\boldsymbol{P}_{t-1}$.
Second, for each neighbor, the feature of the query point, the state of the neighbor and the displacement from the neighbor to the query point are concatenated, which are then processed by a shared fully-connected (FC) layer.
At last, the processed concatenations are reduced to a single representation by pooling.

The PointRNN provides a basic component for point cloud sequence processing. 
Because it may encounter the same exploding and vanishing gradient problems as RNN, we further propose two variants for PointRNN, \ie, Point Gated Recurrent Unit (PointGRU) and Point Long Short-Term Memory (PointLSTM), by combining PointRNN with  GRU and LSTM, respectively. 
We apply PointRNN, PointGRU and PointLSTM to moving point cloud prediction.  
Given the history movements of a point cloud, the goal of this task is to predict the future trajectories of its points.
Predicting how point clouds move in future can help robots and self-driving cars to plan their actions and make decisions.
Moreover, moving point cloud prediction has an innate advantage that it does not require human-annotated supervision. 
Based on PointRNN, PointGRU and PointLSTM, we build up sequence-to-sequence  (seq2seq)~\cite{DBLP:conf/nips/SutskeverVL14} models for moving point cloud prediction.
Experimental results on a synthetic moving MNIST point cloud dataset and two large-scale autonomous driving datasets, \ie, Argoverse~\cite{Chang_2019_CVPR} and nuScenes~\cite{DBLP:journals/corr/abs-1903-11027}, show that PointRNN, PointGRU and PointLSTM produce correct predictions, confirming their ability to process moving point clouds.

\section{Related Work}
\textbf{Static point cloud understanding.} 
The dawn of point cloud has boosted a number of applications, such as object classification, object part segmentation, scene semantic 
segmentation~\cite{DBLP:conf/cvpr/QiSMG17,DBLP:conf/nips/QiYSG17,DBLP:conf/iccv/KlokovL17,DBLP:conf/cvpr/LiCL18,DBLP:conf/nips/LiBSWDC18,DBLP:conf/cvpr/WangYHN18,DBLP:conf/cvpr/GrahamEM18,DBLP:conf/cvpr/SuJSMK0K18,DBLP:journals/cvpr/abs-1904-07601,DBLP:journals/corr/abs-1811-07246,DBLP:journals/corr/abs-1902-09852,DBLP:conf/iccv/abs-1904-08889}, reconstruction~\cite{dai2017scannet,DBLP:conf/aaai/LinKL18,DBLP:conf/cvpr/YuLFCH18} and object detection~\cite{DBLP:conf/cvpr/ChenMWLX17,DBLP:conf/cvpr/ZhouT18,DBLP:conf/cvpr/QiLWSG18,DBLP:journals/corr/abs-1812-04244,DBLP:journals/corr/abs-1812-05784,DBLP:journals/iccv/abs-1904-09664} in 3D space.
Most recently works aim to consume point sets without transforming coordinates to regular 3D voxel grids.
There exists two main challenges for static point cloud processing. 
First, a point cloud is in essence a set of unordered points and invariant to permutations of its points, which necessitates certain symmetrizations in computation.
Second, different with images, point cloud data is irregular. 
Convolutions that capture local structures in images can not be directly applied to such data format.
Different with these existing works, we focus on a new challenge, \ie, to model dynamics in point cloud sequences.

\textbf{RNN variants for spatiotemporal modeling.}
Because the conventional RNN, GRU and LSTM learn from vectors where the representation is highly compact and the spatial structure is heavily damaged, a number of variants are proposed to model spatiotemporal sequences.
For example, convolutional LSTM (ConvLSTM)~\cite{DBLP:conf/nips/ShiCWYWW15} modifies LSTM by taking three-dimensional tensors as input and replacing FC by convolution to capture spatial local structures.
Based on ConvLSTM, spatiotemporal LSTM (ST-LSTM)~\cite{DBLP:conf/nips/WangLWGY17} extracts and memorizes spatial and temporal representations simultaneously by adding a spatiotemporal memory. 
Cubic LSTM (CubicLSTM)~\cite{fan19cubiclstm} extends ConvLSTM by splitting states into temporal states and spatial states, which are respectively generated by independent convolutions. 
However, these methods focus on 2D videos and can not be directly applied to 3D point cloud sequences. 

\section{PointRNN}

In this section, we first review the standard (vanilla) RNN and then describe the proposed PointRNN in detail.

\newcounter{TempEqCnt1}  
\setcounter{TempEqCnt1}{\value{equation}}
\setcounter{equation}{1}                      
\begin{figure*}[h]
\begin{equation}\label{eq:pointrnn}
\small 
\boldsymbol{S}_t = \mathrm{point\mbox{-}rnn}\big((\boldsymbol{P}_t, \boldsymbol{X}_t), (\boldsymbol{P}_{t-1}, \boldsymbol{S}_{t-1}); \boldsymbol{W}, \boldsymbol{b})\big) = \left\{
\underset{j | \boldsymbol{P}_{t-1}^j \in \mathcal{N}(\boldsymbol{P}_t^i)}{\mathrm{pooling}}
\big\{\boldsymbol{W} \cdot \big[\boldsymbol{X}_t^i, \boldsymbol{S}_{t-1}^j,  \boldsymbol{P}_t^i - \boldsymbol{P}_{t-1}^j\big] + \boldsymbol{b}\big\}\right\}_{i \in \{1,\cdots,n\}}
\end{equation}
\vspace{-1.5em}
\end{figure*}
\setcounter{equation}{\value{TempEqCnt1}}

The RNN is a class of deep neural networks which uses its state $\boldsymbol{s}$ to process sequences of inputs.
This allows it to exhibit temporal dynamic behavior.
The RNN relies on a concatenation operation to aggregate the past and the current, which is referred to as the $\mathrm{rnn}$ function in this paper,
\begin{equation}\label{eq:rnn}
\small
\boldsymbol{s}_t = \mathrm{rnn}(\boldsymbol{x}_t, \boldsymbol{s}_{t-1}; \boldsymbol{W}, \boldsymbol{b}) = 
\boldsymbol{W} \cdot [\boldsymbol{x}_t, \boldsymbol{s}_{t-1}] + \boldsymbol{b},
\end{equation}
where $\cdot$ denotes matrix multiplication and $[\cdot, \cdot]$ denotes concatenation.
The $\boldsymbol{W} \in \mathbb{R}^{d' \times (d+d')}$ and $\boldsymbol{b} \in \mathbb{R}^{d'}$ are the parameters of RNN to be learned.
This operation can be implemented by fully-connected (FC) layer in deep neural networks (shown in Figure~\ref{fig:unit}(a)). 
Usually, RNN uses its state $\boldsymbol{s}_t$ as output, \ie, $\boldsymbol{y}_t = \boldsymbol{s}_t$. 

The conventional RNN learns from one-dimensional vectors, limiting its application to process point cloud sequences. 
To keep the spatial structure, we propose to directly take the coordinates $\boldsymbol{P}_t \in \mathbb{R}^{n \times 3}$ and features $\boldsymbol{X}_t \in \mathbb{R}^{n \times d}$ as input.  
Similarly, the state $\boldsymbol{s}_t$ and the output $\boldsymbol{y}_t$ in RNN are extended to $\boldsymbol{S}_t \in \mathbb{R}^{n\times d'}$ and $\boldsymbol{Y}_t \in \mathbb{R}^{n\times d''}$ in PointRNN.
The update for the point states of PointRNN at the $t$-th time step is formulated in Eq.~(\ref{eq:pointrnn}), where $\boldsymbol{W} \in \mathbb{R}^{d' \times (d+d'+3)}$,  $\boldsymbol{b} \in \mathbb{R}^{d'}$ and $\mathcal{N}(\cdot)$ is neighborhood query. 
We refer to this point-based spatiotemporally-local correlation as $\mathrm{point\mbox{-}rnn}$.
Similar to RNN, by default, PointRNN uses $\boldsymbol{S}_t$ as output, \ie, $\boldsymbol{Y}_t = \boldsymbol{S}_t$.

The goal of the $\mathrm{point\mbox{-}rnn}$ function is to aggregate the past and the  current according to point coordinates (shown in Figure~\ref{fig:unit}~(b)).
Given $(\boldsymbol{P}_t, \boldsymbol{X}_t)$ and $(\boldsymbol{P}_{t-1}, \boldsymbol{S}_{t-1})$, $\mathrm{point\mbox{-}rnn}$ merges $\boldsymbol{X}_t$ and $\boldsymbol{S}_{t-1}$ according to $\boldsymbol{P}_t$ and $\boldsymbol{P}_{t-1}$. 
Specifically, for each point in $\boldsymbol{P}_t$, taking the $i$-th point $\boldsymbol{P}_t^i$ for instance, $\mathrm{point\mbox{-}rnn}$ first finds its  neighbors in $\boldsymbol{P}_{t-1}$. 
The neighbors potentially share the same geometry and motion information about the query point $\boldsymbol{P}_t^i$.
Second, for each neighbor, taking the neighbor $\boldsymbol{P}_{t-1}^j$ for instance, the feature $\boldsymbol{X}_t^i$ of the query point, the state $\boldsymbol{S}_{t-1}^j$ of the neighbor, and the displacement $\boldsymbol{P}_t^i - \boldsymbol{P}_{t-1}^j$ from the neighbor to the query point are concatenated, which are subsequently processed by a shared FC layer.
Third, the processed concatenations are pooled to a single representation.
The output of $\mathrm{point\mbox{-}rnn}$, \ie, $\boldsymbol{S}_t$, contains the past and the current information of each point in $\boldsymbol{P}_t$. 

We adopt two methods for neighborhood query. 
The first one directly finds $k$ nearest neighbors for the query point ($k$NN).
The second one works by first finding all points that are within a radius to the query point and then sampling $k$ neighbors from the points (ball query~\cite{DBLP:conf/nips/QiYSG17}). 
Since a point cloud is a set of orderless points, point states or features of two time steps at the same row may correspond to different points. 
Without point coordinates, an independent $\boldsymbol{S}_t$ or $\boldsymbol{Y}_t$ is meaningless.  
Therefore, the point coordinates $\boldsymbol{P}_t$ is added into the state and output of PointRNN. 

The PointRNN provides a prototype that uses RNN to process point cloud sequence.
Each component in PointRNN is necessary.
However, we can design more effective spatiotemporally-local correlation methods to replace $\mathrm{point\mbox{-}rnn}$, which can be further studied in the future.

\section{PointGRU and PointLSTM}
Because PointRNN inherits RNN, it may encounter the same exploding and vanishing gradient problems as RNN. 
Therefore, we propose two variants, which combine PointRNN with GRU and LSTM, respectively, to overcome these problems.
We replace the concatenation operations in GRU with the spatiotemporally-local correlation operations, \ie, $\mathrm{point\mbox{-}rnn}$, forming the PointGRU unit.
The updates for PointGRU at the $t$-th time step are formulated as follows,  
\setcounter{equation}{2} 
\begin{equation}
\footnotesize
\begin{split}
& \boldsymbol{Z}_t  = \sigma\Big(\mathrm{point\mbox{-}rnn}\big((\boldsymbol{P}_t,\boldsymbol{X}_t), (\boldsymbol{P}_{t-1}, \boldsymbol{S}_{t-1}); \boldsymbol{W}_z, \boldsymbol{b}_z\big)\Big), \\
& \boldsymbol{R}_t  = \sigma\Big(\mathrm{point\mbox{-}rnn}\big((\boldsymbol{P}_t,\boldsymbol{X}_t), (\boldsymbol{P}_{t-1}, \boldsymbol{S}_{t-1}); \boldsymbol{W}_r, \boldsymbol{b}_r\big)\Big), \\
& \hat{\boldsymbol{S}}_{t-1}  = \mathrm{point\mbox{-}rnn}\big((\boldsymbol{P}_t, \mathrm{None}), (\boldsymbol{P}_{t-1}, \boldsymbol{S}_{t-1}); \boldsymbol{W}_{\hat s}, \boldsymbol{b}_{\hat s}\big), \\
& \tilde{\boldsymbol{S}}_t = \mathrm{tanh}\big(\boldsymbol{W}_{\tilde s} \cdot [\boldsymbol{X}_t, \boldsymbol{R}_t \odot \hat{\boldsymbol{S}}_{t-1}] + \boldsymbol{b}_{\tilde s}\big), \\
& \boldsymbol{S}_t  = \boldsymbol{Z}_t \odot \hat{\boldsymbol{S}}_{t-1} + (1 -  \boldsymbol{Z}_t) \odot \tilde{\boldsymbol{S}}_t, \\
\end{split}  
\end{equation}
where $\boldsymbol{Z}_t$ is the update gate and $\boldsymbol{R}_t$ is the reset gate.
The $\sigma(\cdot)$ denotes the sigmoid function and $\odot$ denotes the Hadamard product.
Similar to PointRNN, the input of PointGRU is $(\boldsymbol{P}_t, \boldsymbol{X}_t)$, the state is $(\boldsymbol{P}_t, \boldsymbol{S}_t)$ and by default, the output is $(\boldsymbol{P}_t, \boldsymbol{S}_t)$.
Note that, besides the $\mathrm{point\mbox{-}rnn}$ function, another difference between GRU and PointGRU is that, PointGRU has an additional step $\hat{\boldsymbol{S}}_{t-1}$.
The goal of this step is to weight and permute $\boldsymbol{S}_{t-1}$ according to the current input points $\boldsymbol{P}_t$.
Only after this step can we perform Hadamard product between the previous state $\hat{\boldsymbol{S}}_{t-1}$ and the current reset gate $\boldsymbol{R}_t$. 

Similarly, the concatenation operations in LSTM are replaced with the $\mathrm{point\mbox{-}rnn}$ functions, forming the PointLSTM unit.
The updates for PointLSTM at the $t$-th time step are formulated as follows,
\begin{equation}
\footnotesize
\begin{split}
& \boldsymbol{I}_t  = \sigma\Big(\mathrm{point\mbox{-}rnn}\big((\boldsymbol{P}_t,\boldsymbol{X}_t), (\boldsymbol{P}_{t-1}, \boldsymbol{H}_{t-1}); \boldsymbol{W}_i, \boldsymbol{b}_i\big)\Big),\\
& \boldsymbol{F}_t  = \sigma\Big(\mathrm{point\mbox{-}rnn}\big((\boldsymbol{P}_t,\boldsymbol{X}_t), (\boldsymbol{P}_{t-1}, \boldsymbol{H}_{t-1}); \boldsymbol{W}_f, \boldsymbol{b}_f\big)\Big),\\
& \boldsymbol{O}_t  =  \sigma\Big(\mathrm{point\mbox{-}rnn}\big((\boldsymbol{P}_t,\boldsymbol{X}_t), (\boldsymbol{P}_{t-1}, \boldsymbol{H}_{t-1}); \boldsymbol{W}_o, \boldsymbol{b}_o\big)\Big),\\
& \hat{\boldsymbol{C}}_{t-1}  = \mathrm{point\mbox{-}rnn}\big((\boldsymbol{P}_t, \mathrm{None}), (\boldsymbol{P}_{t-1}, \boldsymbol{C}_{t-1}); \boldsymbol{W}_{\hat c}, \boldsymbol{b}_{\hat c}\big),\\
& \tilde{\boldsymbol{C}}_t  = \mathrm{tanh}\Big(\mathrm{point\mbox{-}rnn}\big((\boldsymbol{P}_t,\boldsymbol{X}_t), (\boldsymbol{P}_{t-1}, \boldsymbol{H}_{t-1}); \boldsymbol{W}_{\tilde c}, \boldsymbol{b}_{\tilde c }\big)\Big),\\
& \boldsymbol{C}_t  = \boldsymbol{F}_t \odot \hat{\boldsymbol{C}}_{t-1} + \boldsymbol{I}_t \odot \tilde{\boldsymbol{C}}_t,\\
& \boldsymbol{H}_t  = \boldsymbol{O}_t \odot \mathrm{tanh}(\boldsymbol{C}_t), \\
\end{split}  
\end{equation}
where $\boldsymbol{I}_t$ is the input gate, $\boldsymbol{F}_t$ is the forget gate, $\boldsymbol{O}_t$ is the output gate, $\boldsymbol{C}_t$ is the cell state and $\boldsymbol{H}_t$ is the hidden state.
The input of PointLSTM is $(\boldsymbol{P}_t, \boldsymbol{X}_t)$, the state is $(\boldsymbol{P}_t, \boldsymbol{H}_t, \boldsymbol{C}_t)$ 
and by default, the output is $(\boldsymbol{P}_t, \boldsymbol{H}_t)$. 
Like from GRU to PointGRU, compared with LSTM, PointLSTM has an additional step $\hat{\boldsymbol{C}}_{t-1}$, which weights and permutes $\boldsymbol{C}_{t-1}$ according to the current input points $\boldsymbol{P}_t$.

\section{Moving Point Cloud Prediction}

Moving point clouds provide a large amount of geometric information in scenes as well as profound dynamic changes in motions.
Understanding scenes and imagining motions in 3D space are fundamental abilities for robot and self-driving car platforms.
It is indisputable that an intelligent system, which is able to predict future trajectories of points in a cloud, will have these abilities.
In this paper, we apply PointRNN, PointGRU and PointLSTM to moving point cloud prediction.
Because this task does not require external human-annotated supervisions, models can be trained in an unsupervised manner.

\subsection{Architecture}
\begin{figure*}[t]
\centering
\includegraphics[width=0.995\linewidth]{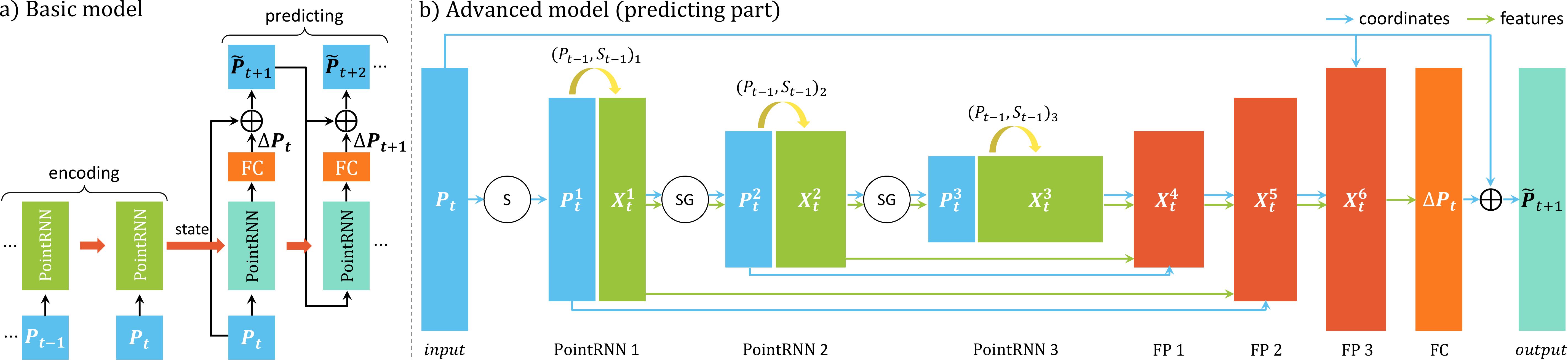}
\footnotesize
\caption{Architectures for moving point cloud prediction. 
a) Basic model (with one PointRNN layer). A PointRNN encodes the given point cloud sequence to a  state, \ie, $(\boldsymbol{P}_t, \boldsymbol{S}_t$), which is then used to initialize the state of a predicting PointRNN. Multiple PointRNN layers can be stacked along the output direction. 
The prediction $\tilde{\boldsymbol{P}}_{t+1}$ is achieved by $\boldsymbol{P}_t+\Delta \boldsymbol{P}_t$.
b) Predicting part of the advanced model (with three PointRNN layers).
Sampling (S) and grouping (G) are used to down-sample points and group features, respectively. PointRNNs aggregate the past and the current, and outputs features for prediction. 
Feature propagation (FP) layers are used to propagate features from subsampled points to the original points.
The fully-connected (FC) layer regresses features to predicted displacements $\Delta \boldsymbol{P}_t$.
}
\label{fig:arch}
\vspace{-1.0em}
\end{figure*}

We design a basic model and an advanced model based on the seq2seq framework.
The basic model (shown in Figure~\ref{fig:arch} (a)) is composed two parts, \ie, one part for encoding the given point cloud sequence and one part for predicting.
Specifically, an encoding PointRNN, PointGRU or PointLSTM unit watches the input point clouds one by one.
After watching the last input $\boldsymbol{P}_t$, its state is used to initialize the state of a  predicting PointRNN, PointGRU or PointLSTM unit.
The predicting unit then takes $\boldsymbol{P}_t$ as input and begins to make predictions.
Rather than directly generating future point coordinates, the proposed models predict displacements $\Delta \boldsymbol{P}_t$ that will happen between the current step and the next step, which can be seen as 3D scene flow~\cite{flownet3d,HPLFlowNet}.

We can stack multiple PointRNN, PointGRU or PointLSTM units to build up a multi-layer structure for hierarchical prediction.
However, because all points are processed in each layer, a major problem of this structure is that it is computationally intensive, especially for high-resolution point sets. 
To alleviate this problem, we propose an advanced model (shown in Figure~\ref{fig:arch} (b)). 
This model borrows two components from PointNet++~\cite{DBLP:conf/nips/QiYSG17}, \ie, 1) a sampling operation and a grouping operation for down-sampling points and their features and 2) a feature propagation layer for up-sampling the representations associated with the intermediate points to the original points.
By this down-up-sampling structure, the advanced model can take advantage of hierarchical learning, while reduce the points to be processed in each layer.

\subsection{Training}
There are two strategies to train recurrent neural networks. 
The first one uses ground truths as inputs during predicting, which is known as \textbf{teacher-forcing} training.
The second one works using predictions generated by the network as inputs (Figure~\ref{fig:arch} (a)), which is referred to as \textbf{free-running} training.
When using the teacher-forcing training, we find that models quickly get stuck in a bad local optima, in which $\Delta \boldsymbol{P}_t$ tends to be $\boldsymbol{0}$ for all inputs.
Therefore, we adopt the free-running training strategy.

Since point clouds are unordered, point-to-point loss functions can not directly apply to compute the difference between the prediction and ground truth.
Loss functions should be invariant to the order of input points. 
In this paper, we adopt Chamfer Distance (CD) and Earth Mover’s Distance (EMD).
The CD between the point set $\boldsymbol{P}$ and $\boldsymbol{P}'$ is defined as follows, 
\begin{equation}\label{eq:chamfer}
\footnotesize
\mathcal{L}_{CD}(\boldsymbol{P}, \boldsymbol{P}') = \sum_{\boldsymbol{p} \in \boldsymbol{P}} \min_{\boldsymbol{p}' \in \boldsymbol{P}'} \|\boldsymbol{p}-\boldsymbol{p}'\|^2 +  \sum_{\boldsymbol{p}' \in \boldsymbol{P}'} \min_{\boldsymbol{p} \in \boldsymbol{P}} \|\boldsymbol{p}'-\boldsymbol{p}\|^2.
\end{equation}
Basically, this loss function is a nearest neighbour distance metric that bidirectionally measures the error in two sets. 
Every point in $\boldsymbol{P}$ is mapped to the nearest point in $\boldsymbol{P}'$, and vice versa. 
The EMD from $\boldsymbol{P}$ to $\boldsymbol{P}'$ is defined as follows,
\begin{equation}\label{eq:emd}
\footnotesize
\mathcal{L}_{EMD}(\boldsymbol{P}, \boldsymbol{P}') = \min_{\phi: \boldsymbol{P} \rightarrow \boldsymbol{P}'} \sum_{\boldsymbol{p} \in \boldsymbol{P}} \|\boldsymbol{p} - \phi(\boldsymbol{p})\|^2,
\end{equation}
where $\phi: \boldsymbol{P} \rightarrow \boldsymbol{P}'$ is a bijection.
The EMD calculates a point-to-point mapping between two point clouds.
The overall loss is as follows,
\begin{equation}\label{eq:loss}
\footnotesize
\mathcal{L}(\boldsymbol{P}, \boldsymbol{P}') = \alpha \mathcal{L}_{CD}(\boldsymbol{P}, \boldsymbol{P}') + \beta \mathcal{L}_{EMD}(\boldsymbol{P}, \boldsymbol{P}'),
\end{equation}
where the hyperparameter $\alpha, \beta \ge 0$.

\section{Experiments}
We conduct experiments on one synthetic moving MNIST point cloud dataset and two large-scale real-world datasets, \ie, Argoverse~\cite{Chang_2019_CVPR} and nuScenes~\cite{DBLP:journals/corr/abs-1903-11027}.
Models are trained for 200k iterations using the Adam~\cite{DBLP:journals/corr/KingmaB14}  optimizer with a learning rate of $10^{-5}$. Gradients are clipped in the range $[-5, 5]$. 
The $\alpha$ and $\beta$ in Eq. (\ref{eq:loss}) are set to 1.0. 
We follow point cloud reconstruction~\cite{DBLP:conf/iclr/AchlioptasDMG18,DBLP:conf/cvpr/YuLFCH18,DBLP:conf/wacv/MandikalR19} to adopt CD and EMD for quantity evaluation.
Model details are listed in Table~\ref{tab:architecture-specs}.
Max pooling is used for our models.
For the two real-world datasets, because they contain too many points to be processed by the basic model, we only evaluate the advanced model.
We randomly synthesize or sample 5,000 sequences from test subsets of these datasets as the test data. 

\begin{table}[t]
\footnotesize
\vspace{0.3em}
\begin{center}
\setlength{\tabcolsep}{3.0pt}
\begin{tabular}{l|cccc|cccc|cccc}
\hline
\multirow{3}{*}{C}                   &   \multicolumn{8}{c|}{Moving MNIST Point Cloud} & \multicolumn{4}{c}{Argoverse \& nuScenes} \\ \cline{2-13}

& \multicolumn{4}{c|}{basic model}      & \multicolumn{4}{c|}{advanced model}  & \multicolumn{4}{c}{advanced model}   \\  \cline{2-13}
    & \#pts & $r$   & $k$   & $c$       & \#pts & $r$   & $k$   & $c$       & \#pts & $r$   & $k$   & $c$   \\ \hline
S   & \multicolumn{4}{c|}{-}            & $n/2$ & 1.0   & 4     & -         & $n/2$ & 0.5   & 8     & -     \\
PU  & $n$   & 4.0   & 8     & 64        & $n/2$ & 4.0   & 12    & 64        & $n/2$ & 1.0   & 24    & 128   \\
SG  & \multicolumn{4}{c|}{-}            & $n/4$ & 2.0   & 4     & -         & $n/4$ & 1.0   & 8     & -     \\
PU  & $n$   & 8.0   & 8     & 128       & $n/4$ & 8.0   & 8     & 128       & $n/4$ & 2.0   & 16    & 256   \\
SG  & \multicolumn{4}{c|}{-}            & $n/8$ & 4.0   & 4     & -         & $n/8$ & 2.0   & 8     & -     \\
PU  & $n$   & 12.0  & 8     & 256       & $n/8$ & 12.0  & 4     & 256       & $n/8$ & 4.0   & 8     & 512   \\
FP  & \multicolumn{4}{c|}{-}            & $n/4$ & -     &-      & 128       & $n/4$ & -     & -     & 256   \\
FP  & \multicolumn{4}{c|}{-}            & $n/2$ & -     &-      & 128       & $n/2$ & -     & -     & 256   \\
FP  & \multicolumn{4}{c|}{-}            & $n$   & -     &-      & 128       & $n$   & -     & -     & 256   \\
FC  & $n$   & -     & -     & 64        & $n$   & -     & -     & 64        & $n$   & -     & -     & 128   \\
FC  & $n$   & -     & -     & 3         & $n$   & -     & -     & 3         & $n$   & -     & -     & 3     \\ \hline
\end{tabular}
\end{center}
\vspace{-1.0em}
\caption{Architecture specs. Each component (C) is described by four attributes, \ie, number of output points (\#pts), search radius ($r$), number of neighbors ($k$) and number of output channels ($c$). S: sampling. G: grouping. PU: PointRNN, PointGRU or PointLSTM unit. FP: feature propagation. FC: fully-connected layer.}
\label{tab:architecture-specs}
\vspace{-1.0em}
\end{table}

\begin{figure*}[t]
\centering
\includegraphics[width=0.995\linewidth]{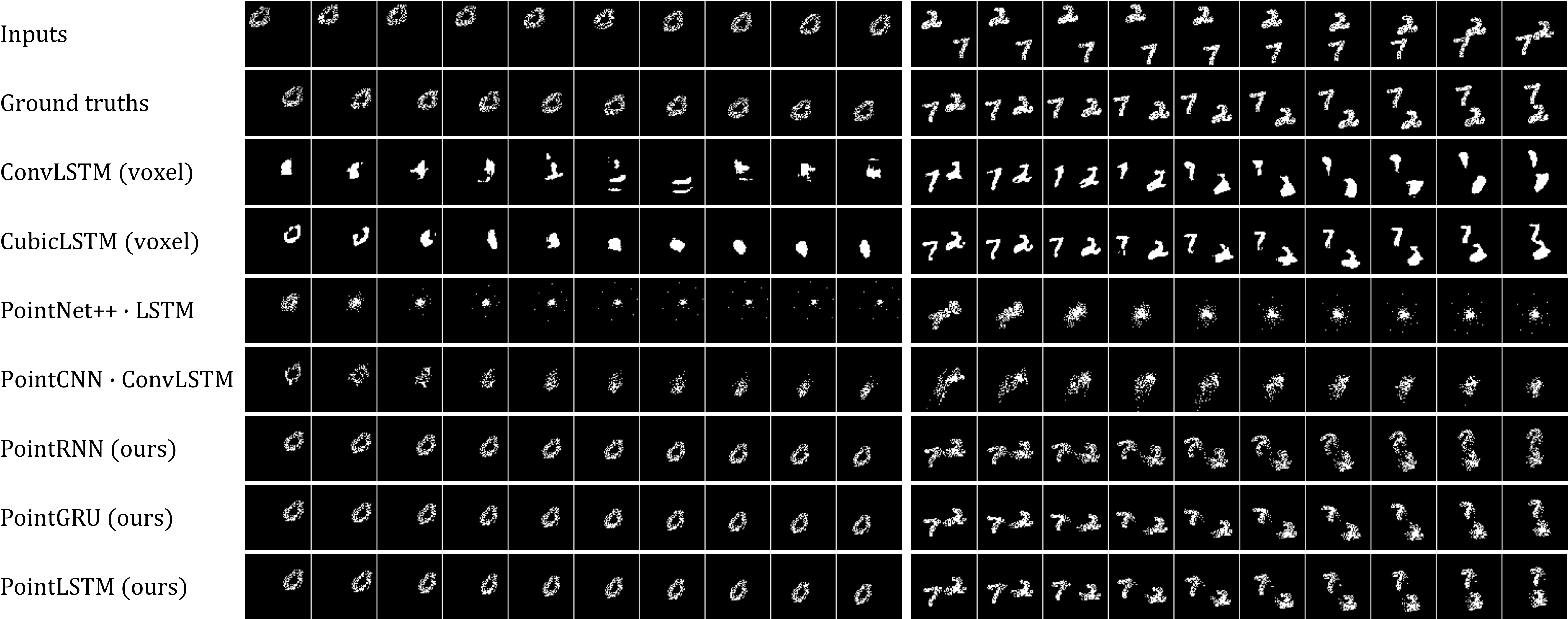}
\footnotesize
\caption{Visualization of moving MNIST point cloud prediction. Left: one moving digit. Right: two moving digits. 
}
\label{fig:main-mmnist}
\end{figure*}

\begin{table*}[t]
\footnotesize
\begin{center}
\setlength{\tabcolsep}{12.25pt}
\begin{tabular}{ll|c|rrr|rrr}
\hline
\multicolumn{2}{l|}{\multirow{2}{*}{Method}}    & \multirow{2}{*}{\#params}     & \multicolumn{3}{c|}{One digit}    & \multicolumn{3}{c}{Two digits}    \\ \cline{4-9}
&   &   & \multicolumn{1}{c}{FLOPs} & \multicolumn{1}{c}{CD} & \multicolumn{1}{c|}{EMD} & \multicolumn{1}{c}{FLOPs} & \multicolumn{1}{c}{CD} & \multicolumn{1}{c}{EMD} \\ \hline 
\multicolumn{2}{l|}{Copy last input}                & -         & \multicolumn{1}{c}{-}     & 262.46    & 15.94     & \multicolumn{1}{c}{-}         & 140.14    & 15.18     \\ 
\multicolumn{2}{l|}{ConvLSTM~\cite{DBLP:conf/nips/ShiCWYWW15} (voxel-based)}    & 2.16      & 345.26    & 58.09     & 8.85      & 345.26    & 13.02     & 5.99      \\ 
\multicolumn{2}{l|}{CubicLSTM~\cite{fan19cubiclstm} (voxel-based)}              & 6.08      & 448.88    & 9.51      & 4.20      & 448.88    & 6.19      & 4.42      \\ 
\multicolumn{2}{l|}{PointNet++ $\cdot$ LSTM}                                    & 2.50      & 1.01      & 175.26    & 15.86     & 1.94      & 100.08    & 15.10      \\ 
\multicolumn{2}{l|}{PointCNN $\cdot$ ConvLSTM}                                  & 2.97      & 1.43      & 15.37     & 5.45      & 2.86      & 49.92     & 10.31      \\ \hline

\multicolumn{1}{c|}{\multirow{3}{*}{\rotatebox[origin=c]{90}{Basic}}}    & PointRNN     & \textbf{0.27}  & 5.32  & 2.65      & 2.79      & 10.65     & 16.08     & 6.62      \\ 
\multicolumn{1}{c|}{}                                                    & PointGRU     & 0.96  & 14.84 & 1.43      & 2.03      & 29.73     & 7.29      & 4.87      \\ 
\multicolumn{1}{c|}{}                                                    & PointLSTM    & 1.22  & 24.74 & 1.40      & 2.00      & 49.55     & 7.28      & 4.82      \\ \hline
\multicolumn{1}{c|}{\multirow{3}{*}{\rotatebox[origin=c]{90}{Advance}}}   & PointRNN     & 0.36  & \textbf{0.77}  & 1.78      & 2.32      & \textbf{1.54}      & 11.44     & 5.87      \\ 
\multicolumn{1}{c|}{}                                                    & PointGRU     & 1.04  & 2.00  & 1.18      & 1.85      & 4.00      & 5.79      & 4.49      \\
\multicolumn{1}{c|}{}                                                    & PointLSTM    & 1.30  & 3.18  & \textbf{1.16} & \textbf{1.78}     & 6.37      & \textbf{5.18} & \textbf{4.21}     \\
\hline
\end{tabular}
\end{center}
\vspace{-1.0em}
\caption{Prediction error (CD and EMD), \#params (million) and FLOPs (billion) on the moving MNIST point cloud dataset.}
\label{tab:mmnist}
\vspace{-0.5em}
\end{table*}

\begin{table}[t]
\footnotesize
\begin{center}
\setlength{\tabcolsep}{10.3pt}
\begin{tabular}{ll|cc|cc}
\hline
\multicolumn{2}{l|}{\multirow{2}{*}{Method}}   & \multicolumn{2}{c|}{One digit}         & \multicolumn{2}{c}{Two digits}  \\ \cline{3-6}
&                                                                   & CD                & EMD                & CD               & EMD  \\ \hline 
\multicolumn{1}{c|}{\multirow{3}{*}{\rotatebox[origin=c]{90}{Basic}}}   & PointRNN      & 5.86              & 3.76               & 22.12            & 7.79 \\
\multicolumn{1}{c|}{}                                                   & PointGRU      & 2.61              & 2.53               & 19.74            & 7.64 \\
\multicolumn{1}{c|}{}                                                   & PointLSTM     & 2.72              & 2.56               & 17.34            & 7.20 \\ \hline
\multicolumn{1}{c|}{\multirow{3}{*}{\rotatebox[origin=c]{90}{Advance}}}  & PointRNN      & 2.25              & 2.53               & 14.54            & 6.42 \\
\multicolumn{1}{c|}{}                                                   & PointGRU      & 1.55              & 2.07               & 8.88            & 5.33 \\
\multicolumn{1}{c|}{}                                                   & PointLSTM     & \textbf{1.43}     & \textbf{1.98}      & \textbf{8.29}    & \textbf{5.13} \\ \hline
\end{tabular}
\end{center}
\vspace{-1.0em}
\caption{Prediction error of PointRNN, PointGRU and PointLSTM with $k$NN on the moving MNIST point cloud dataset.}
\label{tab:mmnist-knn}
\vspace{-1.0em}
\end{table}

We design three baselines for comparison.
\textbf{1) Copy last input.} This method does not make predictions and just copies the last input in the given sequence as outputs.
Models that outperform this method can prove their effectiveness to model point cloud sequence.
\textbf{2) PointNet++ $\cdot$ LSTM.} This method first uses set abstraction (SA) layers from PointNet++~\cite{DBLP:conf/nips/QiYSG17} to encode a point cloud to a global feature. Then, LSTMs take the global features as inputs and output predicted features. At last, like the advanced model, FP layers are used to propagate features to predicted displacements. 
\textbf{3) PointCNN $\cdot$ ConvLSTM}. The PointCNN~\cite{DBLP:conf/nips/LiBSWDC18} provides an $\mathcal{X}$-Conv operator that can weight and permute input points and features before they are processed by a typical convolution. Therefore, PointCNN and ConvLSTM can be combined for point cloud sequence processing by using $\mathcal{X}$-Conv to permute ConvLSTM states. 
We evaluate this combination with the advanced model.

\subsection{Moving MNIST Point Cloud}
Experiments on the synthetic Moving MNIST point cloud dataset can provide some basic understanding of the behavior of the proposed recurrent units.
To synthesize moving MNIST digit sequences, we use a generation process similar to that described in~\cite{DBLP:conf/icml/SrivastavaMS15}.
Each synthetic sequence consists of 20 consecutive point clouds, with 10 for inputs and 10 for predictions. 
Each point cloud contains one or two potentially overlapping handwritten digits moving and bouncing inside a $64 \times 64$ area.
Pixels whose brightness values (ranged from 0 to 255) are less than 16 are removed.
Locations of the remaining pixels are transformed to $(x,y)$ coordinates.
The $z$-coordinate is set to 0 for all points.
We randomly sample 128 points for one digit and 256 points for two digits as input, respectively.
Batch size is set to 32.

Besides the three baselines, we also compare our methods with two video prediction models, \ie, ConvLSTM~\cite{DBLP:conf/nips/ShiCWYWW15} and CubicLSTM~\cite{fan19cubiclstm}.
Essentially, the Moving MNIST point cloud dataset is 2D.
Digit point clouds can be first voxelized to images and then apply video-based methods to process them.
Specifically, a pixel in the voxelization image is set to 1 if there exists a point at the position. Otherwise, the pixel is set to 0.
In this way, a 2D point cloud sequence is converted to a video.  
For training, we use the binary cross entropy loss to optimize each output pixel.
For evaluation, because the number of output points is usually not consistent with that of input points, we collect points whose brightness are in top 128 (for one digit) or top 256 (for two digits) as the output point cloud.

\begin{table}[t]
\footnotesize
\vspace{-0.2em}
\begin{center}
\setlength{\tabcolsep}{2.8pt}
\begin{tabular}{ll|c|ccc|ccc}
\hline
\multicolumn{2}{l|}{\multirow{2}{*}{Method}}   & \multirow{2}{*}{\#params} & \multicolumn{3}{c|}{One digit}         & \multicolumn{3}{c}{Two digits}  \\ \cline{4-9}
&       &                                                   & FLOPs         & CD                & EMD          & FLOPs       & CD               & EMD  \\ \hline 
\multicolumn{1}{c|}{\multirow{3}{*}{\rotatebox[origin=c]{90}{1 layer}}}   & PointRNN  & \textbf{0.025} & \textbf{0.176}   & 8.23          & 4.35      & \textbf{0.354}       & 29.92            & 8.30 \\
\multicolumn{1}{c|}{}                                                     & PointGRU  & 0.051 & 0.454   & \textbf{3.72}          & \textbf{2.99}      & 0.912     & 19.41            & \textbf{6.99} \\
\multicolumn{1}{c|}{}                                                     & PointLSTM & 0.060 & 0.710   & 4.34          & 3.17      & 1.427    & \textbf{18.37}            & 7.14 \\ \hline
\multicolumn{1}{c|}{\multirow{3}{*}{\rotatebox[origin=c]{90}{2 layers}}}  & PointRNN  & \textbf{0.109} & \textbf{0.485}   & 2.77          & 2.80      & \textbf{0.972}      & 16.51            & 6.62 \\
\multicolumn{1}{c|}{}                                                     & PointGRU  & 0.268 & 1.224    & \textbf{1.48}          & \textbf{2.05}      & 2.453    & 8.22            & 5.13 \\
\multicolumn{1}{c|}{}                                                     & PointLSTM & 0.328 & 1.962    & 1.50          & 2.06      & 3.931     & \textbf{7.82}            & \textbf{4.96} \\ \hline
\end{tabular}
\end{center}
\vspace{-1.0em}
\caption{Performance of advanced models (ball query) with different numbers of layers on the moving MNIST point cloud dataset.}
\label{tab:mmnist-layer}
\vspace{-1.0em}
\end{table}

Experimental results are listed in Table~\ref{tab:mmnist} (our methods are with ball query).  
The $k$NN results are listed in Table~\ref{tab:mmnist-knn}.
The performance of advanced models (ball query) with different numbers of layers is reported in Table~\ref{tab:mmnist-layer}. 
Two prediction examples are visualized in Figure~\ref{fig:main-mmnist} (our models are with the advanced architecture and ball query).

PointRNN, PointGRU and PointLSTM are superior to other methods.
For example, in Table~\ref{tab:mmnist}, the CD of advanced PointLSTM on one-digit prediction is 1.16, less than ConvLSTM by 56.93, CubicLSTM by 8.35, PointNet++ $\cdot$ LSTM by 174.1 and PointCNN $\cdot$ ConvLSTM by 14.21.

\begin{table}[t]
\footnotesize
\begin{center}
\setlength{\tabcolsep}{2.2pt}
\begin{tabular}{l|cc|cc|ccc}
\hline

\multirow{2}{*}{Dataset}    &   \multicolumn{2}{c|}{trainval}    & \multicolumn{2}{c|}{test}  &  \multirow{2}{*}{frequency}  & \multirow{2}{*}{\#pts/pc}  & \multirow{2}{*}{range}   \\ \cline{2-5}
                            & \#logs    & \#pcs     & \#logs    & \#pcs     &      &        \\ \hline    
Argoverse~\cite{Chang_2019_CVPR} & 89  & 18,211 & 24 & 4,189 & 10Hz & 90,549 & 200m \\
nuScenes~\cite{DBLP:journals/corr/abs-1903-11027} & 68  & 297,737 & 15 & 52,423 & 20Hz & 34,722  & 70m\\ \hline
\end{tabular}
\end{center}
\vspace{-1.0em}
\caption{Details of the Argoverse and nuScenes datasets. \#pcs: number of point clouds, \#pts/pc: number of points per point cloud on average.}
\label{tab:dataset}
\end{table}

\begin{table}[t]
\footnotesize
\vspace{-0.5em}
\begin{center}
\setlength{\tabcolsep}{1.44pt}
\begin{tabular}{l|cc|cc|cc}
\hline

\multirow{2}{*}{Method}        &   \multicolumn{1}{c}{\multirow{2}{*}{\#params}}  &   \multicolumn{1}{c|}{\multirow{2}{*}{FLOPs}}  &   \multicolumn{2}{c|}{Argoverse}  &   \multicolumn{2}{c}{nuScenes}   \\ \cline{4-7}
         &             &             & CD      & EMD           & CD        & EMD \\ \hline 
Copy last input                 & - & - & 0.5812 & 1.0667 & 0.0794 & 0.3961  \\ 
PointNet++ $\cdot$ LSTM         & \ \ 9.98 & 26.54 & 0.3826 & 1.0011 & 0.0716 & 0.3953  \\ 
PointCNN $\cdot$ ConvLSTM       & 12.04 & 24.17 & 0.3457 & 0.9659 & 0.0683 & 0.3874  \\  \hline
PointRNN (ours)                 & \ \ \textbf{1.42} & \textbf{22.40} & \textbf{0.2789} & 0.8964 & \textbf{0.0619} & 0.3750  \\ 

PointGRU (ours)                 & \ \ 4.15 & 59.49 & 0.2994 & 0.9084 & 0.0620 & \textbf{0.3738}  \\ 
PointLSTM (ours)                & \ \ 5.18 & 98.49 & 0.2966 & \textbf{0.8892} & 0.0624 & 0.3745  \\ \hline
\end{tabular}
\end{center}
\vspace{-1.0em}
\caption{Prediction error (CD and EMD), \#params (million) and FLOPs (billion) on Argoverse and nuScenes.}
\label{tab:argo-nu-ball}
\vspace{-1.0em}
\end{table}

Compared with the basic architecture, the advanced architecture significantly reduces computation. 
For example, in Table~\ref{tab:mmnist}, for the one-digit prediction, the FLOPs of the basic PointGRU is 14.84 billion and the FLOPs of the advanced PointGRU is only 2.00 billion.

The advanced architecture also achieves lower prediction error than the basic architecture. 
For example, in Table~\ref{tab:mmnist}, for the two-digit prediction, the CD of the basic PointRNN is 16.08 and  the CD of the advanced PointRNN is 11.44.


The ball query is superior to the $k$NN. 
For example, when applying the advanced PointLSTM model to the two-digit prediction, the EMD of the ball query is 4.21 (in Table~\ref{tab:mmnist}) and the EMD of the $k$NN is 5.13 (in Table~\ref{tab:mmnist-knn}).

Hierarchical structure by stacking multiple layers can effectively improve accuracy.
For example, on the one-digit prediction, the CDs of the advanced PointGRU with one, two and three layers are 3.72, 1.48 and 1.18, respectively. 

Learning via global features fails to make predictions. 
In Table~\ref{tab:mmnist}, although the PointNet++ $\cdot$ LSTM combination obtains lower CDs than the copy-last-input method, their EMDs are similar.
In Figure~\ref{fig:main-mmnist}, PointNet++ $\cdot$ LSTM fails to predict neither appearance nor motion.

Point-based models consume less FLOPs than voxel-based models. 
Taking the two-digit prediction for instance, the basic PointLSTM model consumes only 49.55 billion FLOPs, while the voxel-based ConvLSTM consumes 345.26 billion FLOPs. 
The reason is that, the FLOPs of point-based models only depends on the number of input points,  while voxel-based models have to process the entire space no matter how many points are actually in the space. 

Voxel-based models are not sensitive to sparse point clouds.
According to Table~\ref{tab:mmnist} and Figure~\ref{fig:main-mmnist}, the predictions of ConvLSTM and CubicLSTM on one moving digit are worse than those on two moving digits, which is counterintuitive. 
One possible reason is that, for one-digit prediction, there is only $128/(64\times64)=3.125\%$  point area, which is too sparse for voxel-based methods. 

\begin{figure}[t]
\centering
\includegraphics[width=0.995\linewidth]{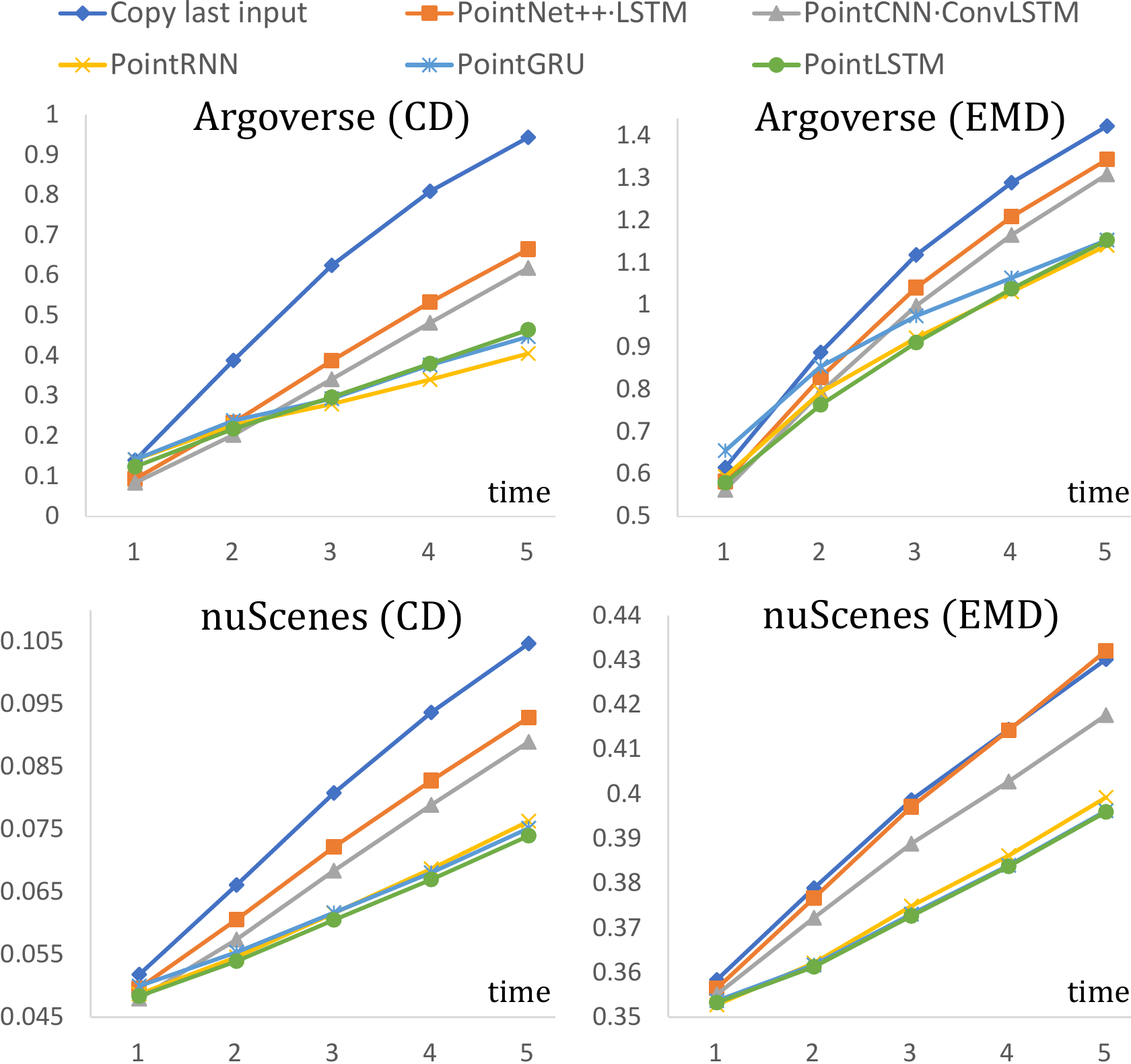}
\footnotesize
\caption{Comparison of different methods over time.
}
\label{fig:chart}
\end{figure}

\begin{table}[t]
\footnotesize
\begin{center}
\setlength{\tabcolsep}{11pt}
\begin{tabular}{l|cc|cc}
\hline

\multirow{2}{*}{Method}         &   \multicolumn{2}{c|}{Argoverse}  &   \multicolumn{2}{c}{nuScenes}   \\ \cline{2-5}
                         & CD      & EMD           & CD        & EMD \\ \hline 
PointRNN                 & \textbf{0.2541} & \textbf{0.8743} & 0.0627 & 0.3739  \\ 
PointGRU                 & 0.2922 & 0.9054 & \textbf{0.0610} & \textbf{0.3711}  \\ 
PointLSTM                & 0.2890 & 0.8856 & 0.0619 & 0.3730  \\ \hline
\end{tabular}
\end{center}
\vspace{-1.0em}
\caption{Prediction error of PointRNN, PointGRU and PointLSTM with $k$NN on Argoverse and nuScenes.}
\label{tab:argo-nu-knn}
\vspace{-1.5em}
\end{table}

\subsection{Argoverse and nuScenes}
Argoverse~\cite{Chang_2019_CVPR} and nuScenes~\cite{DBLP:journals/corr/abs-1903-11027} are two large-scale autonomous driving datasets. 
The Argoverse data is collected by a fleet of autonomous vehicles in Pittsburgh (86km) and Miami (204km).
The nuScenes data is recorded in Boston and Singapore, with 15h of driving data (242km travelled at an average of 16km/h).
These datasets are collected by multiple sensors, including LiDAR, RADAR, camera, \etc.
In this paper, we only use the data from LiDAR sensors, without any human-annotated supervision, for moving point cloud prediction.
Details about Argoverse and nuScenes are listed in Table~\ref{tab:dataset}.

\begin{figure*}[t]
\centering
\includegraphics[width=0.995\linewidth]{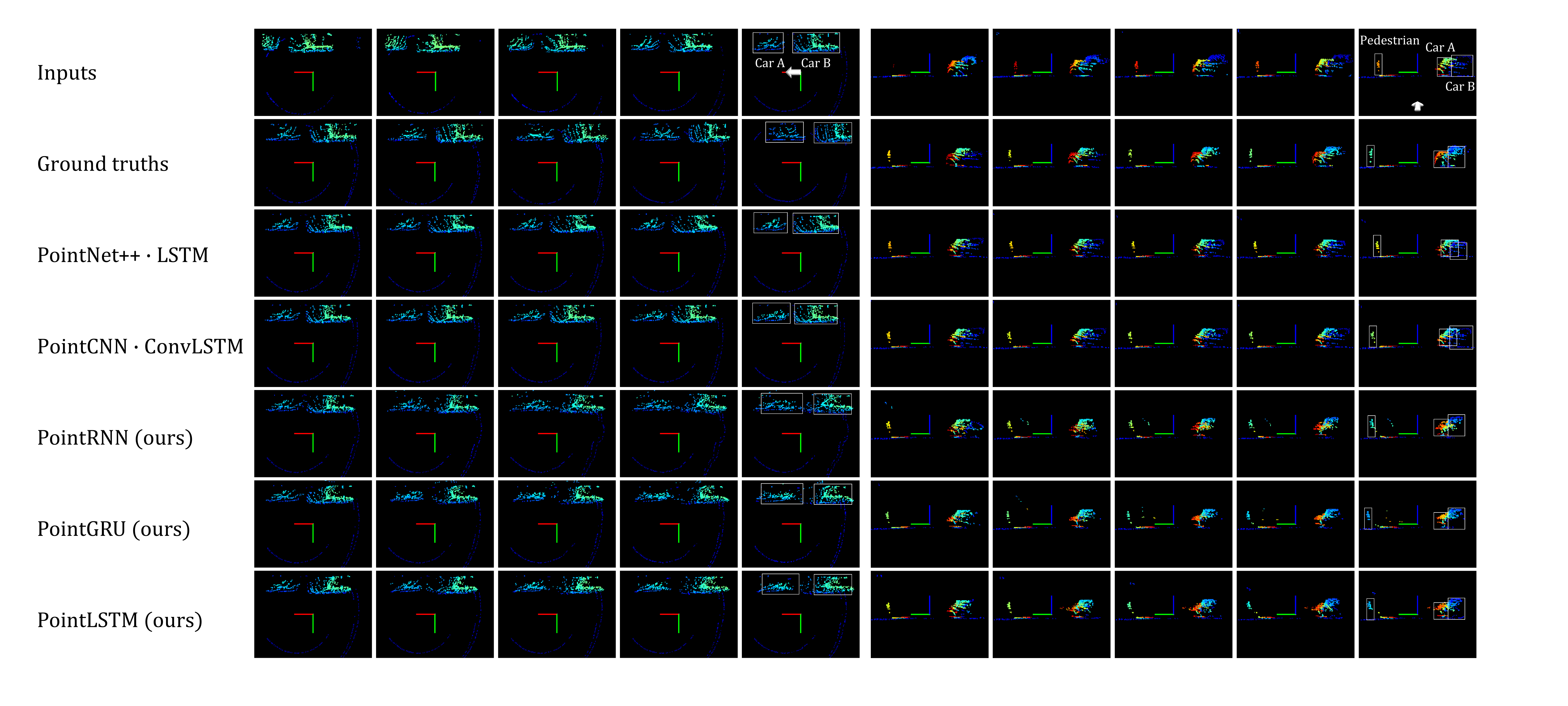}
\footnotesize
\caption{Visualization of moving point cloud prediction on Argoverse. 
Left: an example of bird's-eye view (color encodes height).
Right: an example of worm's-eye view (color encodes depth).
Moving objects are marked with bounding boxes at the last time step.
}
\label{fig:argo-nu}
\vspace{-1.0em}
\end{figure*}

Predicting moving point clouds on real-world driving datasets is considerably challenging.
Content of a long driving point cloud sequence may change dramatically.
Because we can not predict what are not provided in the given inputs, models are asked to make short-term prediction on the driving datasets.
Each driving log is considered as a continuous point cloud sequence.
We randomly choice 10 successive point clouds from a driving log for training, with 5 for inputs and 5 for predictions. 
Since Argoverse and nuScenes are high-resolution, using all points requires considerable computation and running memory.
Therefore, for each cloud, we only use the points whose coordinates are in the range $[-5m, 5m]$.
Then, we randomly sample 1,024 points from the range as inputs and ground truths.
Batch size is set to 4.

Experimental results are listed in Table~\ref{tab:argo-nu-ball} (our methodsare with ball query). 
The result details over time are illustrated in Figure~\ref{fig:chart}. 
The $k$NN results are listed in Table~\ref{tab:argo-nu-knn}. 

PointRNN, PointGRU and PointLSTM are superior to PointNet++ $\cdot$ LSTM and PointCNN $\cdot$ ConvLSTM.
For example, in Table~\ref{tab:argo-nu-ball}, the CD of advanced PointRNN on Argoverse is 0.2789, less than PointNet++ $\cdot$ LSTM by 0.1037 and PointCNN $\cdot$ ConvLSTM by 0.0668.
In Figure~\ref{fig:chart}, our models achieve lower prediction error than PointNet++ $\cdot$ LSTM and PointCNN $\cdot$ ConvLSTM at most time steps.

The $k$NN and ball query achieve similar accuracy on Argoverse and nuScenes.
For example, when applying the advanced PointLSTM model to Argoverse, the EMD of the ball query is 0.8892 (in Table~\ref{tab:argo-nu-ball}) and the EMD of the $k$NN is 0.8856 (in Table~\ref{tab:argo-nu-knn}).

We visualize two prediction examples in Figure~\ref{fig:argo-nu} (our models are with the ball query). 
For the first example, PointNet++ $\cdot$ LSTM and PointCNN $\cdot$ ConvLSTM fail to predict car A or car B to move backward, while PointRNN, PointGRU and PointLSTM can predict the movement of car B correctly and the movement of car A with a little error at the rear. 
For the second example, PointNet++ $\cdot$ LSTM fails to predict the movements of the pedestrian or the two cars. 
The PointCNN $\cdot$ ConvLSTM model can predict the pedestrian to a certain degree, but fails to predict the cars.
PointRNN, PointGRU and PointLSTM can predict the movements of both the pedestrian and the cars.

\begin{figure}[t]
\centering
\vspace{0.25em}
\includegraphics[width=0.995\linewidth]{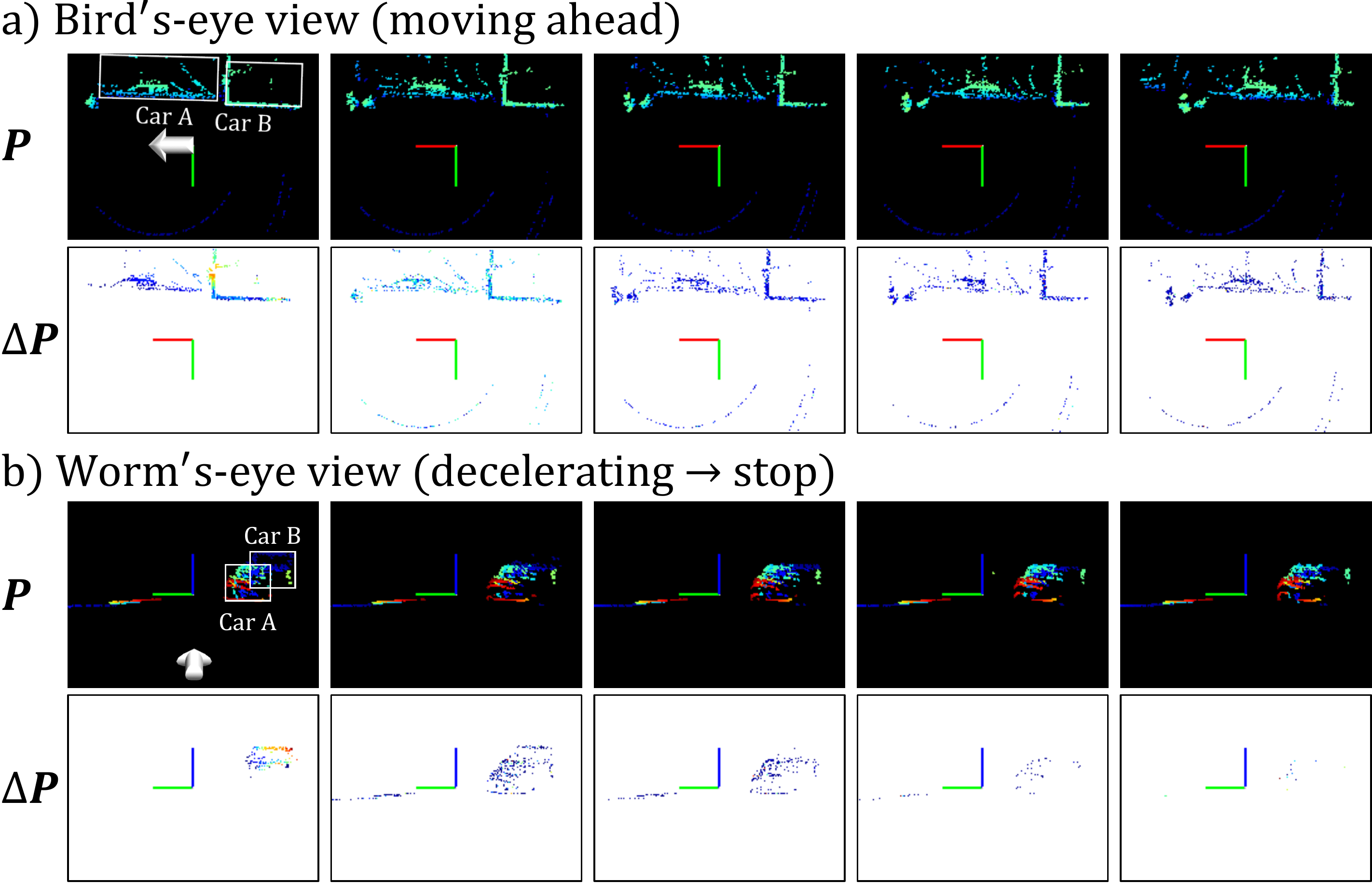}
\footnotesize
\caption{Visualization of predicted scene flow ($\Delta \boldsymbol{P}$). 
The color of scene flow encodes flow magnitude.
a) When the LiDAR is uniformly moving, the model generates similar scene flow. 
b) When the LiDAR stops moving, the scene flow vanishes.
}
\label{fig:flow}
\vspace{-1.0em}
\end{figure}

We also visualize the predicted scene flow ($\Delta \boldsymbol{P}$) by the advanced PointLSTM model (ball query) in Figure~\ref{fig:flow}.
The point whose flow magnitude $||\Delta \boldsymbol{p}||$ is less than $0.01m$ is removed. 
The model outputs reasonable scene flow.
This suggests a promising method to learn 3D scene flow from point clouds in an unsupervised manner.

\section{Conclusion}
This paper proposes a PointRNN, PointGRU and PointLSTM for point cloud sequence processing. 
Experimental results on moving point cloud prediction demonstrate their ability to model point cloud sequences.
The proposed units have the potential for other temporal-related applications, such as 3D  action recognition based on point cloud and sequential scene semantic segmentation. 
More effective spatiotemporally-local correlation methods can be studied to improve PointRNN, PointGRU and PointLSTM.

{\small
\bibliographystyle{ieee_fullname}
\bibliography{egbib}
}


\newpage
\newcommand{\hbAppendixPrefix}{}
\renewcommand{\thesection}{\hbAppendixPrefix\arabic{section}}
\setcounter{section}{0}
\renewcommand{\thefigure}{\hbAppendixPrefix\arabic{figure}}
\setcounter{figure}{0}
\renewcommand{\thetable}{\hbAppendixPrefix\arabic{table}} 
\setcounter{table}{0}
\renewcommand{\theequation}{\hbAppendixPrefix\arabic{equation}} 
\setcounter{equation}{0}

\onecolumn
\begin{center}
\centering
\Large  
\textbf{PointRNN: Point Recurrent Neural Network for Moving Point Cloud Processing}   \textsc{Supplementary Material}
\end{center}
\vspace{9.0 mm}

\section{PointGRU and PointLSTM}
In this section, we provide detailed breakdowns of PointGRU and PointLSTM in the main paper by comparing them with the conventional GRU and LSTM, respectively.

\subsection{PointGRU}
\begin{wrapfigure}{r}{0.4\textwidth}
    \vspace{-2.75em}
    \begin{center}
        \subfigure[]{
            \includegraphics[width=0.18\textwidth]{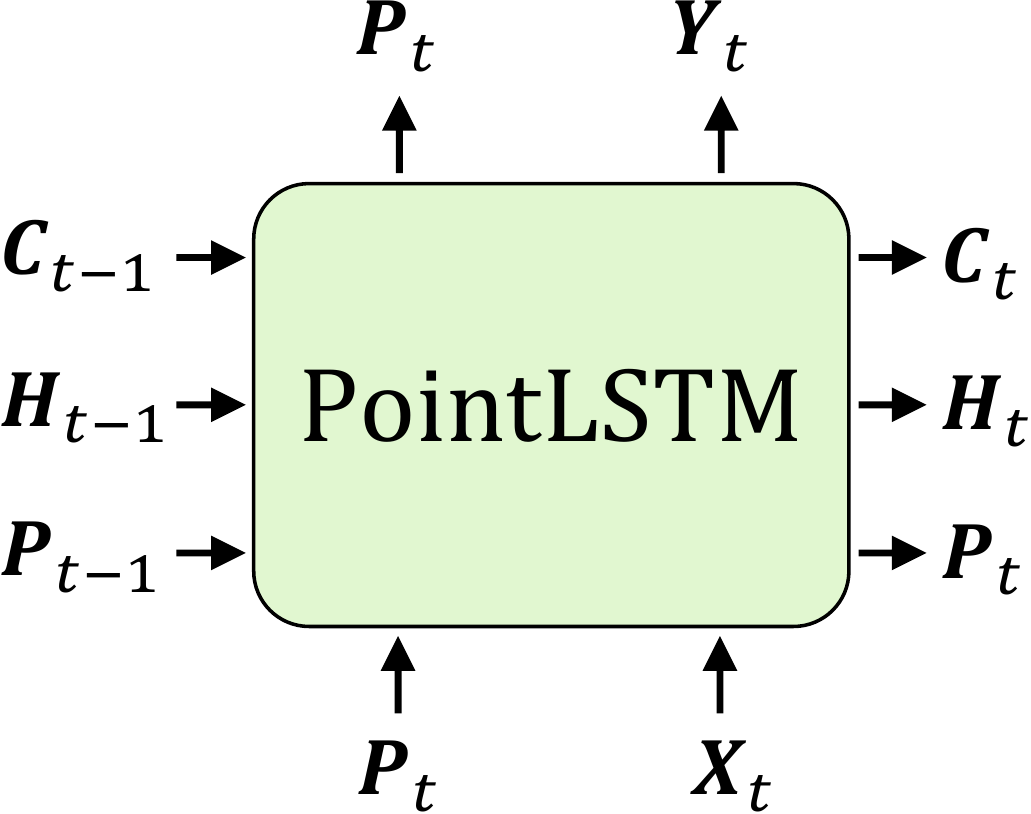}
            \label{fig:structure-pointgru}
        } 
        \hfill
        \subfigure[]{
            \includegraphics[width=0.18\textwidth]{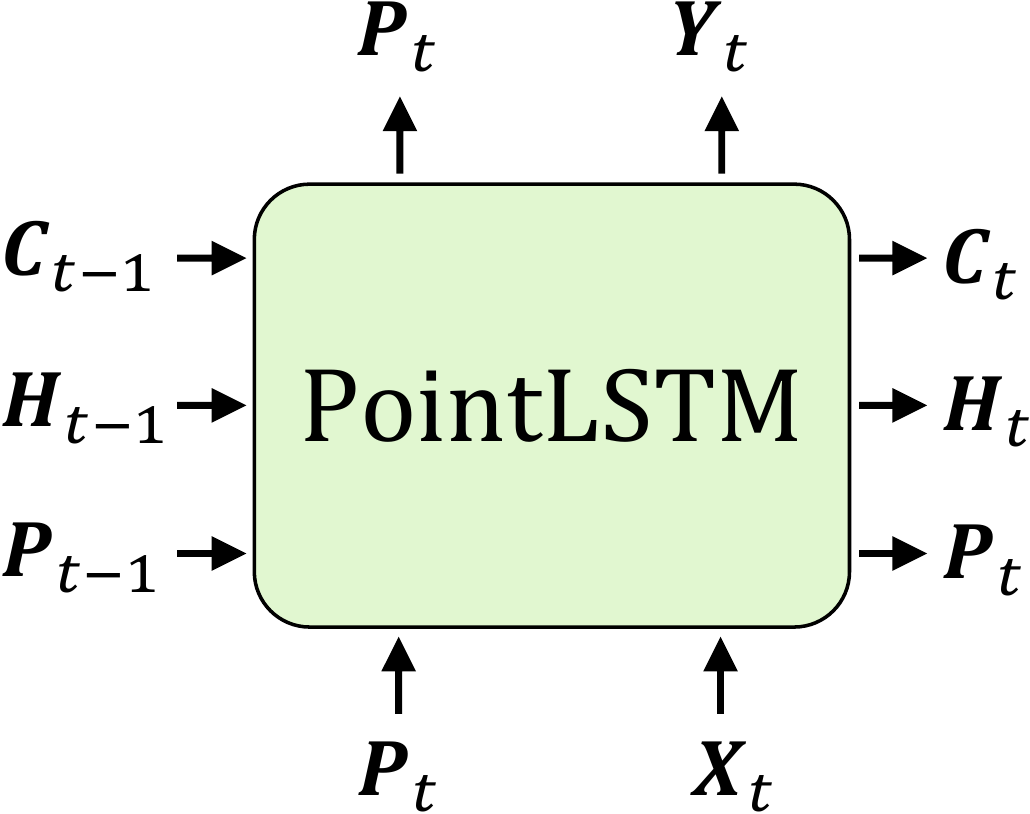}
            \label{fig:structure-pointlstm}
        }
  \end{center}
  \vspace{-1.3em}
  \caption{PointGRU and PointLSTM. By default, the $\boldsymbol{Y}_t$ of PointGRU is set to $\boldsymbol{S}_t$ and the $\boldsymbol{Y}_t$ of PointLSTM is set to $\boldsymbol{H}_t$.}
\end{wrapfigure}
To solve the exploding and vanishing gradient problems of the standard RNN, GRU introduces a update gate $\boldsymbol{z}_t \in \mathbb{R}^{d'}$ and a reset gate $\boldsymbol{r}_t \in \mathbb{R}^{d'}$ to decide what information should be passed to the output. 
The update gate helps the model to determine how much of the past information (from previous time steps) needs to be passed along to the future. 
The reset gate is used to decide how much of the past information to forget.
The current state $\tilde{\boldsymbol{s}}_t \in \mathbb{R}^{d'}$ uses the reset gate to store the relevant information from the past.
The final state $\boldsymbol{s}_t \in \mathbb{R}^{d'}$ uses the update gate to determine what to collect from the current state $\tilde{\boldsymbol{s}}_t$ and what from the previous state $\boldsymbol{s}_{t-1}$.

\begin{table*}[h]
\footnotesize
\vspace{0.5em}
\begin{center}
\setlength{\tabcolsep}{8.25pt}
\begin{tabular}{l|l|l}
\hline
Gate/State       &          \multicolumn{1}{c|}{GRU}                                &   \multicolumn{1}{c}{PointGRU}   \\ \hline 
update gate  & $\boldsymbol{z}_t = \sigma\big(\mathrm{rnn}(\boldsymbol{x}_t, \boldsymbol{s}_{t-1}; \boldsymbol{W}_z, \boldsymbol{b}_z)\big)$ & $\boldsymbol{Z}_t  = \sigma\Big(\mathrm{point\mbox{-}rnn}\big((\boldsymbol{P}_t,\boldsymbol{X}_t), (\boldsymbol{P}_{t-1}, \boldsymbol{S}_{t-1}); \boldsymbol{W}_z, \boldsymbol{b}_z\big)\Big)$ \\
reset gate  & $\boldsymbol{r}_t = \sigma\big(\mathrm{rnn}(\boldsymbol{x}_t, \boldsymbol{s}_{t-1}; \boldsymbol{W}_r, \boldsymbol{b}_r)\big)$ & $\boldsymbol{R}_t  = \sigma\Big(\mathrm{point\mbox{-}rnn}\big((\boldsymbol{P}_t,\boldsymbol{X}_t), (\boldsymbol{P}_{t-1}, \boldsymbol{S}_{t-1}); \boldsymbol{W}_r, \boldsymbol{b}_r\big)\Big)$ \\ \hline
\multirow{3}{*}{state}  & \multicolumn{1}{c|}{-} & $\hat{\boldsymbol{S}}_{t-1}  = \mathrm{point\mbox{-}rnn}\big((\boldsymbol{P}_t, \mathrm{None}), (\boldsymbol{P}_{t-1}, \boldsymbol{S}_{t-1}); \boldsymbol{W}_{\hat s}, \boldsymbol{b}_{\hat s}\big)$ \\
 & $\tilde{\boldsymbol{s}}_t = \mathrm{tanh}\big(\boldsymbol{W}_s \cdot [\boldsymbol{x}_t, \boldsymbol{r}_t \odot \boldsymbol{s}_{t-1}] + \boldsymbol{b}_s\big)$ & 
 $\tilde{\boldsymbol{S}}_t = \mathrm{tanh}\big(\boldsymbol{W}_{\tilde s} \cdot [\boldsymbol{X}_t, \boldsymbol{R}_t \odot \hat{\boldsymbol{S}}_{t-1}] + \boldsymbol{b}_{\tilde s}\big)$ \\
 & $\boldsymbol{s}_t = \boldsymbol{z}_t \odot \boldsymbol{s}_{t-1} + (1 - \boldsymbol{z}_t) \odot \tilde{\boldsymbol{s}}_t$ & $\boldsymbol{S}_t  = \boldsymbol{Z}_t \odot \hat{\boldsymbol{S}}_{t-1} + (1 -  \boldsymbol{Z}_t) \odot \tilde{\boldsymbol{S}}_t$ \\ \hline
\end{tabular}
\end{center}
\vspace{-1.0em}
\caption{Comparison between GRU and PointGRU.}
\label{tab:gru-pointgru}
\vspace{-1.0em}
\end{table*}

\begin{table*}[h]
\footnotesize
\begin{center}
\setlength{\tabcolsep}{7.5pt}
\begin{tabular}{l|l|l}
\hline
Gate/State       &          \multicolumn{1}{c|}{LSTM}                                &   \multicolumn{1}{c}{PointLSTM}   \\ \hline 
input gate  & $\boldsymbol{i}_t = \sigma\big(\mathrm{rnn}(\boldsymbol{x}_t, \boldsymbol{h}_{t-1}; \boldsymbol{W}_i, \boldsymbol{b}_i)\big)$ & $\boldsymbol{I}_t  = \sigma\Big(\mathrm{point\mbox{-}rnn}\big((\boldsymbol{P}_t,\boldsymbol{X}_t), (\boldsymbol{P}_{t-1}, \boldsymbol{H}_{t-1}); \boldsymbol{W}_i, \boldsymbol{b}_i\big)\Big)$ \\
forget gate  & $\boldsymbol{f}_t = \sigma\big(\mathrm{rnn}(\boldsymbol{x}_t, \boldsymbol{h}_{t-1}; \boldsymbol{W}_f, \boldsymbol{b}_f)\big)$ & $\boldsymbol{F}_t  = \sigma\Big(\mathrm{point\mbox{-}rnn}\big((\boldsymbol{P}_t,\boldsymbol{X}_t), (\boldsymbol{P}_{t-1}, \boldsymbol{H}_{t-1}); \boldsymbol{W}_f, \boldsymbol{b}_f\big)\Big)$ \\
output gate  & $\boldsymbol{o}_t = \sigma\big(\mathrm{rnn}(\boldsymbol{x}_t, \boldsymbol{h}_{t-1}; \boldsymbol{W}_o, \boldsymbol{b}_o)\big)$ & $\boldsymbol{O}_t  =  \sigma\Big(\mathrm{point\mbox{-}rnn}\big((\boldsymbol{P}_t,\boldsymbol{X}_t), (\boldsymbol{P}_{t-1}, \boldsymbol{H}_{t-1}); \boldsymbol{W}_o, \boldsymbol{b}_o\big)\Big)$ \\ \hline
\multirow{3}{*}{cell state}  & \multicolumn{1}{c|}{-} & $\hat{\boldsymbol{C}}_{t-1}  = \mathrm{point\mbox{-}rnn}\big((\boldsymbol{P}_t, \mathrm{None}), (\boldsymbol{P}_{t-1}, \boldsymbol{C}_{t-1}); \boldsymbol{W}_{\hat c}, \boldsymbol{b}_{\hat c}\big)$ \\
 & $\tilde{\boldsymbol{c}}_t = \mathrm{tanh}\big(\mathrm{rnn}(\boldsymbol{x}_t, \boldsymbol{h}_{t-1}; \boldsymbol{W}_c, \boldsymbol{b}_c)\big)$ & $\tilde{\boldsymbol{C}}_t  = \mathrm{tanh}\Big(\mathrm{point\mbox{-}rnn}\big((\boldsymbol{P}_t,\boldsymbol{X}_t), (\boldsymbol{P}_{t-1}, \boldsymbol{H}_{t-1}); \boldsymbol{W}_{\tilde c}, \boldsymbol{b}_{\tilde c }\big)\Big)$ \\
 & $\boldsymbol{c}_t = \boldsymbol{f}_t \odot \boldsymbol{c}_{t-1} + \boldsymbol{i}_t \odot \tilde{\boldsymbol{c}}_t$ & $\boldsymbol{C}_t  = \boldsymbol{F}_t \odot \hat{\boldsymbol{C}}_{t-1} + \boldsymbol{I}_t \odot \tilde{\boldsymbol{C}}_t$ \\ \hline
hidden state & $\boldsymbol{h}_t  = \boldsymbol{o}_t \odot \mathrm{tanh}(\boldsymbol{c}_t)$ & $\boldsymbol{H}_t  = \boldsymbol{O}_t \odot \mathrm{tanh}(\boldsymbol{C}_t)$ \\ \hline
\end{tabular}
\end{center}
\vspace{-1.0em}
\caption{Comparison between LSTM and PointLSTM.}
\label{tab:lstm-pointlstm}

\end{table*}

The update gate $\boldsymbol{z}_t$ and the reset gate $\boldsymbol{r}_t$ of GRU are calculated by the $\mathrm{rnn}$ function in the main paper, which aggregates the previous state $\boldsymbol{s}_{t-1}$ and the current input $\boldsymbol{x}_t \in \mathbb{R}^d$ with concatenation. 
By default, GRU uses its state $\boldsymbol{s}_t$ as output $\boldsymbol{y}_t$. 
To apply GRU to point cloud, we replace $\mathrm{rnn}$ in the reset and upset gates with the proposed $\mathrm{point\mbox{-}rnn}$ function, forming the PointGRU unit.
The comparison of update steps between GRU and PointGRU are shown in Table~\ref{tab:gru-pointgru}. 
Because point clouds are unordered, PointGRU leverages an additional step $\hat{\boldsymbol{S}}_{t-1}$ to weight and permute $\boldsymbol{S}_{t-1}$ according to the current input points $\boldsymbol{P}_t$. 
Similar to PointRNN, the input of PointGRU is $(\boldsymbol{P}_t, \boldsymbol{X}_t)$, the state is $(\boldsymbol{P}_t, \boldsymbol{S}_t)$ and the output is $(\boldsymbol{P}_t, \boldsymbol{Y}_t)$ (shown in Figure~\ref{fig:structure-pointgru}). By default, $\boldsymbol{Y}_t$ is set to $\boldsymbol{S}_t$. 

\subsection{PointLSTM}
The conventional LSTM unit is composed of a cell state $\boldsymbol{c}_t \in \mathbb{R}^{d'}$, a hidden state a cell state $\boldsymbol{h}_t \in \mathbb{R}^{d'}$, an input gate $\boldsymbol{i}_t \in \mathbb{R}^{d'}$, an output gate $\boldsymbol{o}_t \in \mathbb{R}^{d'}$ and a forget gate $\boldsymbol{f}_t \in \mathbb{R}^{d'}$. 
The cell state acts as an accumulator of the sequence or the temporal information over time.
Specifically, the current input $\boldsymbol{x}_t$ will be integrated to $\boldsymbol{c}_t$ if the input gate $\boldsymbol{i}_t$ is activated.
Meanwhile, the previous cell state $\boldsymbol{c}_{t-1}$ may be forgotten if the forget gate $\boldsymbol{f}_t$ turns on. 
Whether $\boldsymbol{c}_t$ will be propagated to the hidden state $\boldsymbol{h}_t \in \mathbb{R}^{d'}$ is controlled by the output gate $\boldsymbol{o}_t$.
By default, LSTM uses its hidden state $\boldsymbol{h}_t$ as output $\boldsymbol{y}_t$.

Similar to the gates in GRU, the cell state $\tilde{\boldsymbol{c}}_t$, input gate $\boldsymbol{i}_t$, output gate $\boldsymbol{o}_t$ and forget gate $\boldsymbol{f}_t$ of LSTM are calculated by the $\mathrm{rnn}$ function. 
We replace $\mathrm{rnn}$ in LSTM with the $\mathrm{point\mbox{-}rnn}$ function, forming the PointLSTM.
The comparison of update steps between LSTM and PointLSTM are shown in Table~\ref{tab:lstm-pointlstm}. 
Like from GRU to PointGRU, PointLSTM has an additional step $\hat{\boldsymbol{C}}_{t-1}$ to weight and permute $\boldsymbol{C}_{t-1}$ according to the current input points $\boldsymbol{P}_t$. 
The input of PointLSTM is $(\boldsymbol{P}_t, \boldsymbol{X}_t)$, the state is $(\boldsymbol{P}_t, \boldsymbol{H}_t, \boldsymbol{C}_t)$ and the output is $(\boldsymbol{P}_t, \boldsymbol{Y}_t)$ (shown in  Figure~\ref{fig:structure-pointlstm}).
By default, $\boldsymbol{Y}_t$ is set to $\boldsymbol{H}_t$.

\section{More Experiments}
In this section we provide more experimental results to validate and analyze the proposed PointRNN, PointGRU and PointLSTM.

\subsection{Prediction on Argoverse and nuScenes with Different Ball Query Radiuses}

In the main paper, the search radiuses of the architecture for Argoverse and nuScenes are fixed as $S(0.5)\rightarrow PU(1.0)\rightarrow SG(1.0)\rightarrow PU(2.0)\rightarrow SG(2.0)\rightarrow PU(4.0)$. 
In this section, we evaluate another two radius settings, to explore the impact of ball query radius. Note that,  the number of neighbor (\ie, $k$) in each component is not changed.
\begin{itemize}
    \vspace{-0.25em}
    \setlength\itemsep{-0.25em}
    \item Setting 1: $S(0.05)\rightarrow PU(0.1)\rightarrow SG(0.1)\rightarrow PU(0.2)\rightarrow SG(0.2)\rightarrow PU(0.4)$.
    \item Setting 2: $S(0.15)\rightarrow PU(0.3)\rightarrow SG(0.3)\rightarrow PU(0.6)\rightarrow SG(0.6)\rightarrow PU(1.2)$.
    \item Setting 3: $S(0.25)\rightarrow PU(0.5)\rightarrow SG(0.5)\rightarrow PU(1.0)\rightarrow SG(1.0)\rightarrow PU(2.0)$.
    \vspace{-0.25em}
\end{itemize}

The results are listed in Table~\ref{tab:argo-nu-radius}. 
The radius of ball query have an obvious impact on the prediction accuracy. 
For example, the CD of PointRNN with setting 1 on Argoverse is 0.2875, while that with setting 2 is 0.3001.
This suggests that, for ball query, it can effectively reduce prediction error by carefully choosing search radiuses.

\begin{table}[h]
\footnotesize
\begin{center}
\setlength{\tabcolsep}{13.25pt}
\begin{tabular}{ll|cc|cc}
\hline
\multicolumn{2}{l|}{\multirow{2}{*}{Method}}  &\multicolumn{2}{c|}{Argoverse}  & \multicolumn{2}{c}{nuScenes}   \\ \cline{3-6}
&                          												& CD           & EMD           & CD        & EMD       \\ \hline 
\multicolumn{1}{c|}{\multirow{3}{*}{Setting 1}} 
 													& PointRNN         	& 0.2875       & 0.8906        & 0.0625    & 0.3750  \\ 
\multicolumn{1}{c|}{} 								& PointGRU          & 0.3292       & 0.9834        & 0.0620    & 0.3753  \\ 
\multicolumn{1}{c|}{} 								& PointLSTM         & 0.3556  	   & 0.9610        & 0.0617    & 0.3743  \\ \hline
\multicolumn{1}{c|}{\multirow{3}{*}{Setting 2}} 
													& PointRNN          & 0.3001       & 0.8924        & 0.0610    & 0.3724	 \\
\multicolumn{1}{c|}{} 								& PointGRU          & 0.3084       & 0.9803        & 0.0623    & 0.3719	 \\
\multicolumn{1}{c|}{} 								& PointLSTM         & 0.3036       & 0.9248        & 0.0598    & 0.3727	 \\ \hline
\multicolumn{1}{c|}{\multirow{3}{*}{Setting 3}} 
													& PointRNN          & 0.2798       & 0.8893        & 0.0610    & 0.3719	 \\
\multicolumn{1}{c|}{} 								& PointGRU          & 0.2839       & 0.9388        & 0.0606    & 0.3718	 \\
\multicolumn{1}{c|}{} 								& PointLSTM         & 0.3370       & 0.9590        & 0.0607    & 0.3727	 \\ \hline
\end{tabular}
\end{center}
\vspace{-1.0em}
\caption{Prediction error with different ball query radiuses on Argoverse and nuScenes.}
\label{tab:argo-nu-radius}
\vspace{-0.5em}
\end{table}


\subsection{Average pooling}
In the main paper, PointRNNs, PointGRUs and PointLSTMs are with max pooling. In this section, we investigate average pooling on the moving MNIST point cloud dataset. 
The results are reported in Table~\ref{tab:mmnist-average}. 
Compared with the moving MNIST point cloud results in the main paper, max pooling is better than average pooling. 

\begin{table}[h]
\footnotesize
\begin{center}
\setlength{\tabcolsep}{12.5pt}
\begin{tabular}{ll|cc|cc|cc|cc}
\hline
\multicolumn{2}{l|}{\multirow{3}{*}{Method}}   & \multicolumn{4}{c|}{$k$NN} & \multicolumn{4}{c}{Ball query} \\ \cline{3-10}
& & \multicolumn{2}{c|}{One digit}         & \multicolumn{2}{c|}{Two digits}  & \multicolumn{2}{c|}{One digit}         & \multicolumn{2}{c}{Two digits} \\ \cline{3-10}
&                                                                                       & CD            & EMD           & CD            & EMD           & CD            & EMD           & CD            & EMD   \\ \hline 
\multicolumn{1}{c|}{\multirow{3}{*}{\rotatebox[origin=c]{90}{Basic}}}   & PointRNN      & 10.81         & 4.82          & 25.89         & 8.58          & 4.85          & 3.54          & 23.04         & 7.45  \\
\multicolumn{1}{c|}{}                                                   & PointGRU      & 4.82          & 3.36          & 20.42         & 7.85          & 2.44          & 2.50          & 13.10         & 6.05  \\
\multicolumn{1}{c|}{}                                                   & PointLSTM     & 5.14          & 3.41          & 28.34         & 9.69          & 2.58          & 2.54          & 13.63         & 6.19  \\ \hline
\multicolumn{1}{c|}{\multirow{3}{*}{\rotatebox[origin=c]{90}{Advance}}} & PointRNN      & 4.04          & 3.16          & 16.86         & 6.72          & 2.62          & 2.73          & 15.50         & 6.50  \\
\multicolumn{1}{c|}{}                                                   & PointGRU      & \textbf{2.70} & 2.62          & \textbf{12.50}& \textbf{6.27} & \textbf{1.51} & \textbf{2.06} & \textbf{8.79} & \textbf{5.27}  \\
\multicolumn{1}{c|}{}                                                   & PointLSTM     & 2.75          & \textbf{2.58} & 14.13         & 6.58          & 1.65          & 2.14          & 18.82         & 5.56  \\ \hline
\end{tabular}
\end{center}
\vspace{-1.0em}
\caption{Prediction error of PointRNN, PointGRU and PointLSTM with average pooling on moving MNIST point cloud.}
\label{tab:mmnist-average}
\vspace{-0.5em}
\end{table}

\subsection{More Visualization Examples}
In this section, we provide more visualization examples on the moving MNIST point cloud, Argoverse and nuScenes datasets.
Our methods are with the ball query.

\begin{figure*}[h]
\centering
\includegraphics[width=0.995\linewidth]{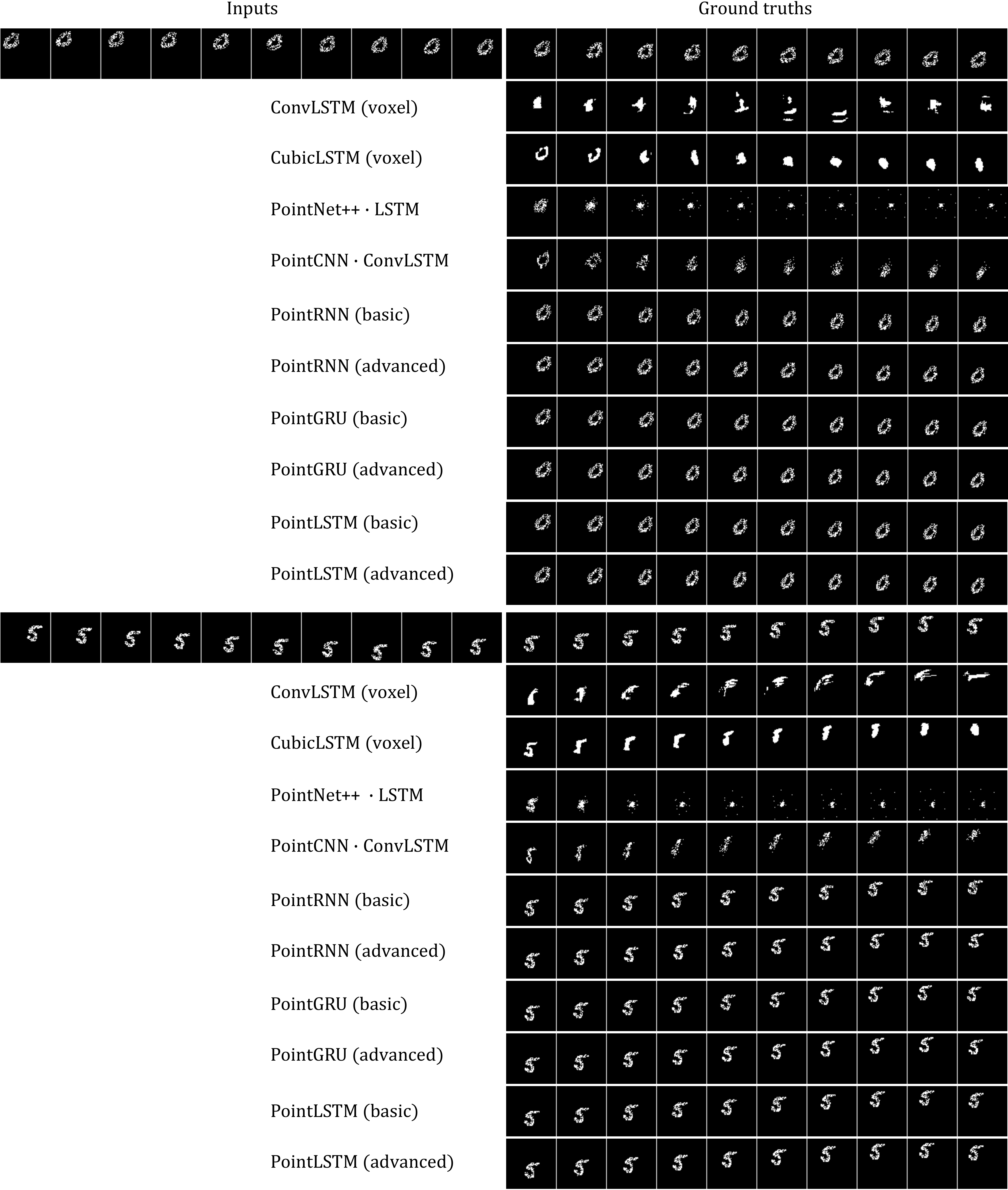}
\footnotesize
\smallskip
\caption{Visualization of moving MNIST point cloud prediction with one digit. 
For PointRNN, PointGRU and PointLSTM, both the basic model and the advanced model make correct appearances and motions.  
}
\label{fig:mmnist-1}
\end{figure*}

\begin{figure*}[h]
\centering
\includegraphics[width=0.995\linewidth]{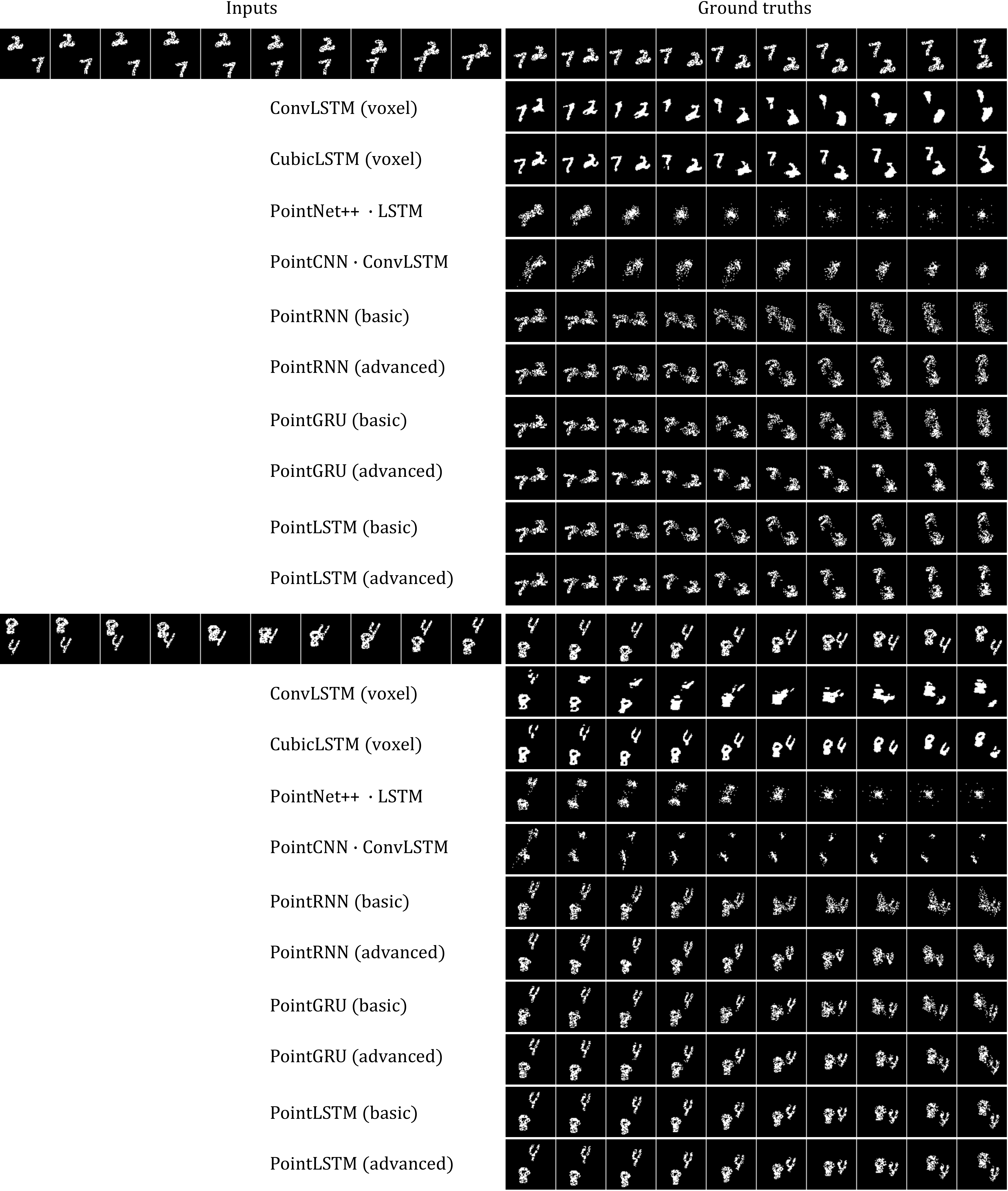}
\footnotesize
\smallskip
\caption{Visualization of moving MNIST point cloud prediction with two digits.
For PointRNN, PointGRU and PointLSTM, the advanced model generates clearer digits than the basic model.}
\label{fig:mmnist-2}
\end{figure*}

\begin{figure*}[t]
\label{sup-argo}
\centering
\subfigure[Bird's-eye view (color encodes height).]{
        \includegraphics[width=0.99\linewidth]{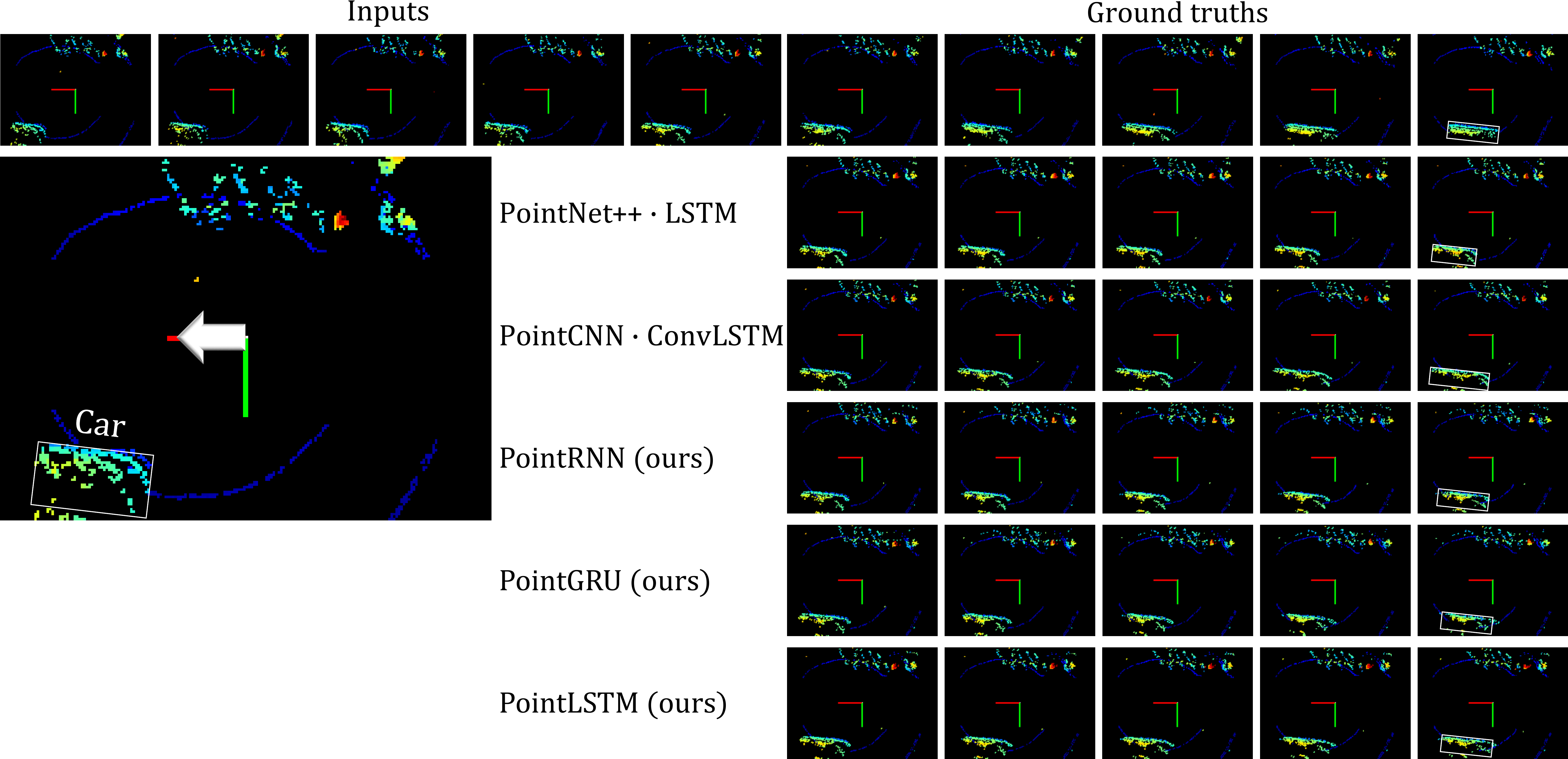}
}
\smallskip
\subfigure[Worm's-eye view (color encodes depth).]{
        \includegraphics[width=0.99\linewidth]{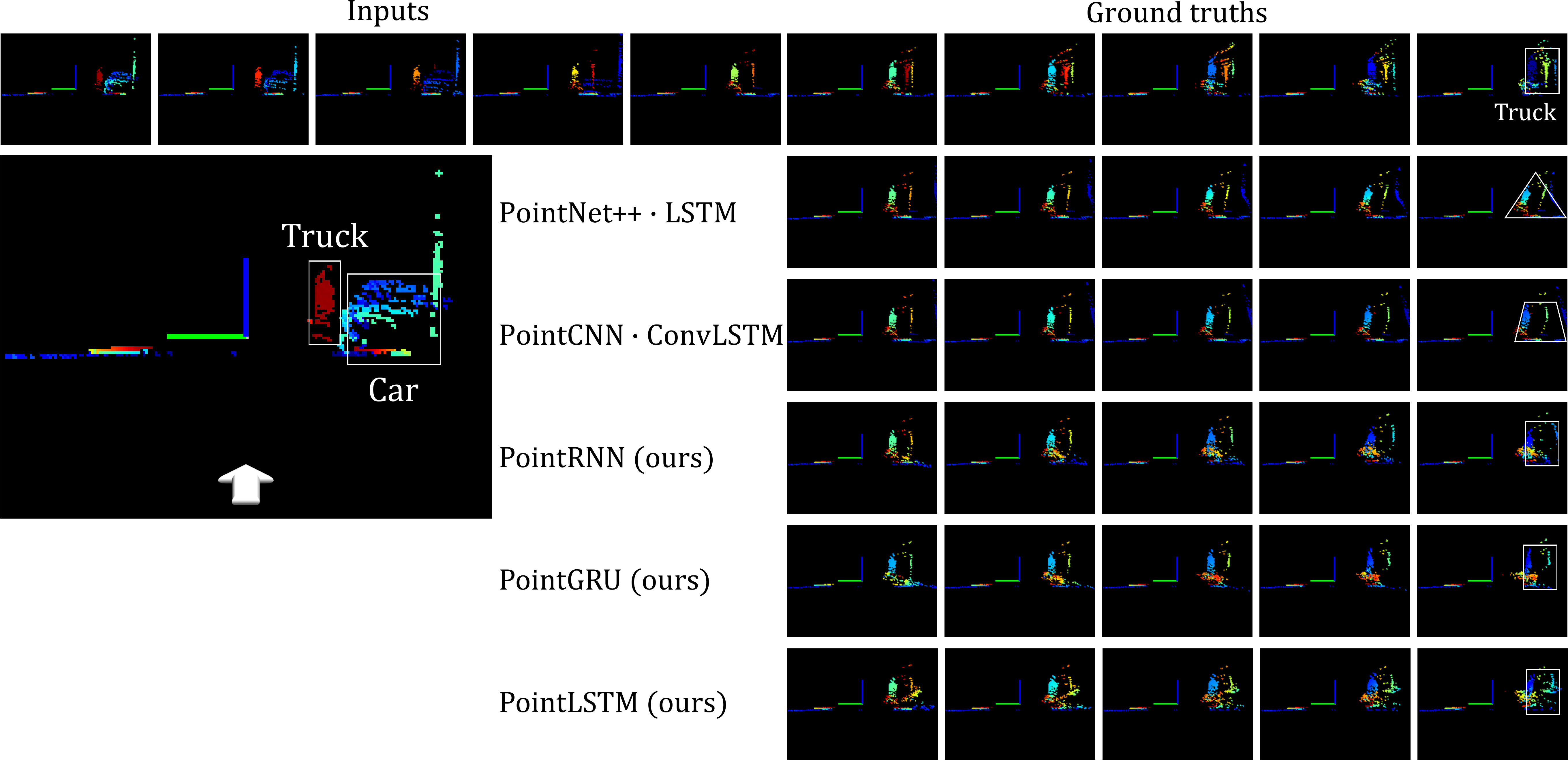}
}
\smallskip
\caption{Visualization of moving point cloud prediction on Argoverse. 
(a) In the bird's-eye view example, two cars are moving backward. 
The PointRNN, PointGRU and PointLSTM correctly predict them to move backward.  
(b) In the worm's-eye view example, a truck and a car are moving backward. 
At last, the car moves out of the field of visualization.
The PointRNN, PointGRU and PointLSTM correctly predict the truck and the car to move backward. 
At the last time step, predictions of the truck made by PointNet++ $\cdot$ LSTM and
PointCNN $\cdot$ ConvLSTM are severely distorted.   
}
\end{figure*}

\begin{figure*}[t]
\label{sup-nu}
\centering
\subfigure[Bird's-eye view (color encodes height).]{
        \includegraphics[width=0.99\linewidth]{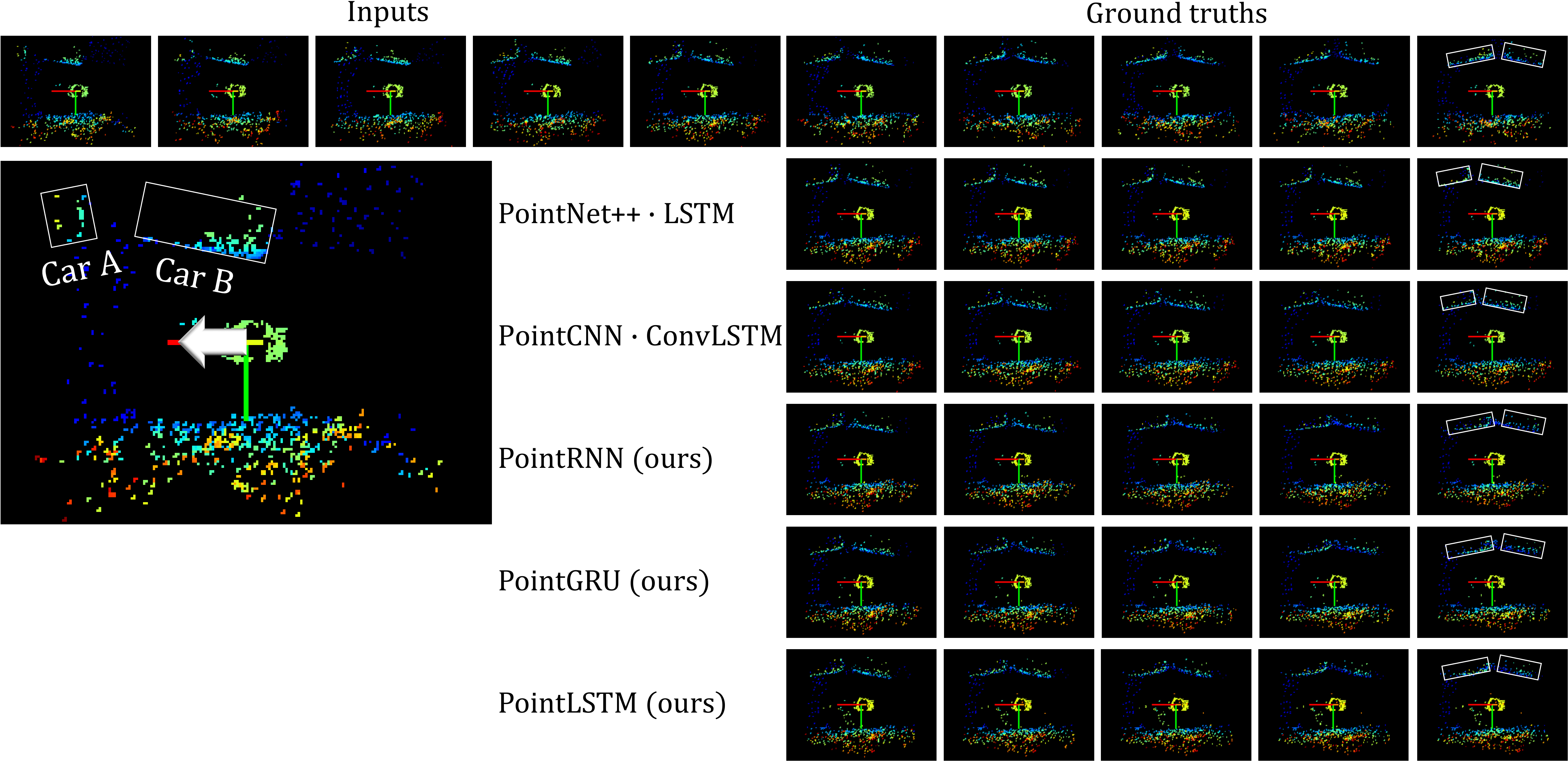}
}
\smallskip
\subfigure[Worm's-eye view (color encodes depth).]{
        \includegraphics[width=0.99\linewidth]{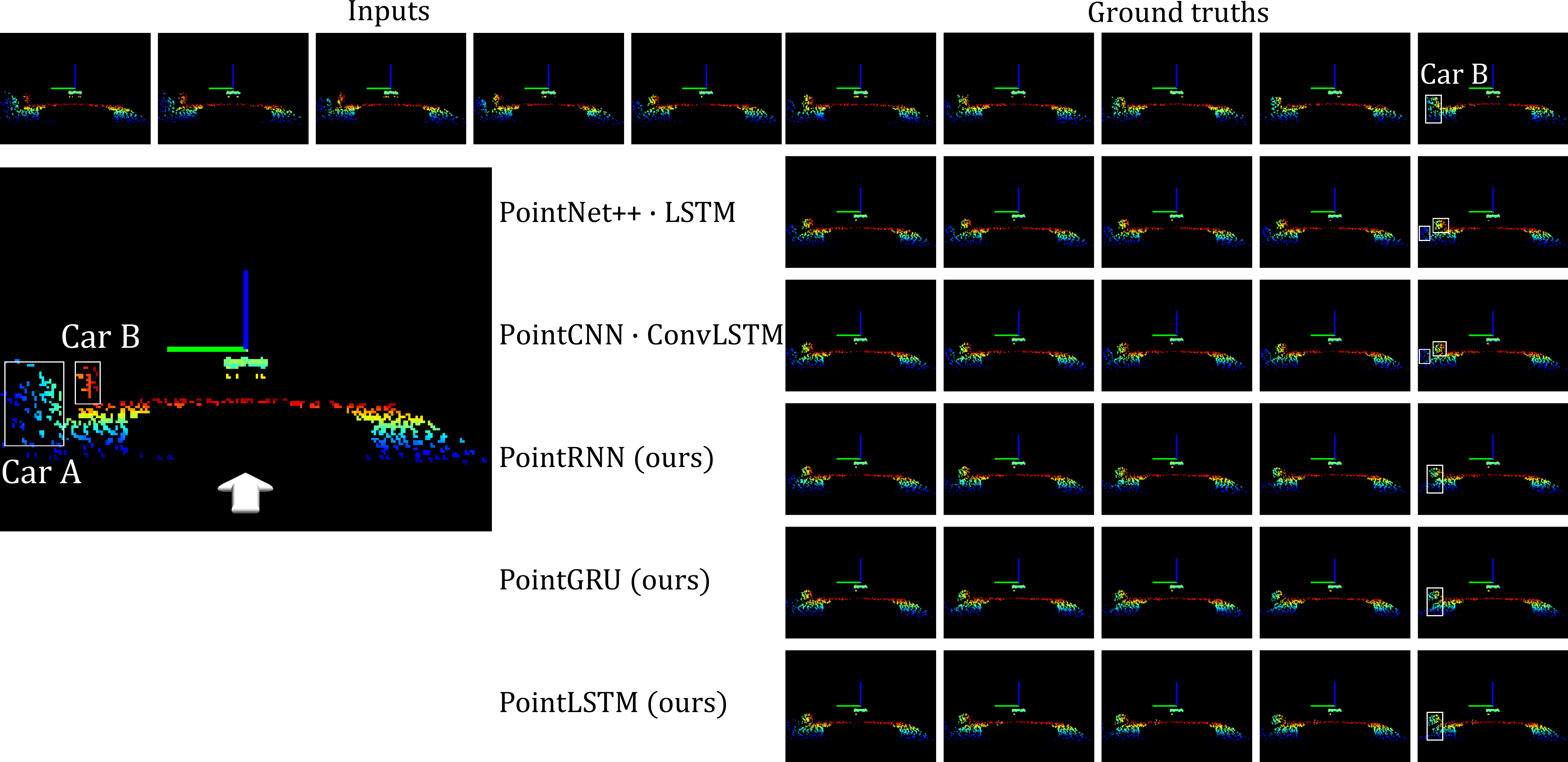}
}
\smallskip
\caption{Visualization of moving point cloud prediction on nuScenes. 
(a) In the bird's-eye view example, two cars are moving backward. 
The PointRNN, PointGRU and PointLSTM correctly predict them to move backward.  
In the worm's-eye view example, there are also two cars which are moving backward. 
(b) In each example, there are also two cars that are moving backward. 
At last, car A moves out of the field of visualization. 
The PointRNN, PointGRU and PointLSTM correctly predict car B to move backward and car A to gradually move out of the field of visualization.
}
\end{figure*}

\end{document}